\documentclass{article}

\usepackage[nonatbib, preprint]{neurips_2020}

\usepackage{comment}
\usepackage{makecell}
\usepackage{colortbl}
\usepackage[utf8]{inputenc}
\usepackage{microtype}
\usepackage{graphicx}
\usepackage{subfigure}
\usepackage{booktabs}
\usepackage{hyperref}
\usepackage{multirow}
\usepackage{todonotes}
\usepackage{todonotes}
\usepackage{caption}
\usepackage{xspace}
\usepackage{xcolor,pifont}
\usepackage[T1]{fontenc}
\definecolor{darkspringgreen}{rgb}{0.09, 0.45, 0.27}

\usepackage{tikz}
\usepackage{pgfplots}
\usepackage[backend=biber,style=alphabetic,natbib=true]{biblatex}
\newcommand{\origds}{\textsc{Original}\xspace}
\newcommand{\mrds}{\textsc{Mixed-Rand}\xspace}
\newcommand{\obgtds}{\textsc{Only-BG-T}\xspace}
\newcommand{\obgds}{\textsc{Only-BG}\xspace}
\newcommand{\obgbds}{\textsc{Only-BG-B}\xspace}
\newcommand{\oads}{\textsc{IN-9L}\xspace}
\newcommand{\mnds}{\textsc{Mixed-Next}\xspace}
\newcommand{\msds}{\textsc{Mixed-Same}\xspace}
\newcommand{\ofgds}{\textsc{Only-FG}\xspace}
\newcommand{\nfgds}{\textsc{No-FG}\xspace}
\newcommand{\mixedds}{\textsc{Mixed}\xspace}
\newcommand{\bggap}{\textsc{BG-Gap}\xspace}

\newcommand{\dsname}{ImageNet-9\xspace}
\newcommand{\dsnamebal}{IN-9LB\xspace}
\newcommand{\dsn}{IN-9\xspace}
\newcommand*\colorcheck[1]{%
  \expandafter\newcommand\csname #1check\endcsname{\textcolor{#1}{\ding{51}}}%
}
\newcommand*\colorx[1]{%
  \expandafter\newcommand\csname #1x\endcsname{\textcolor{#1}{\ding{55}}}%
}

\usepackage[normalem]{ulem}
\usepackage{xcolor}

\newcommand{\scrcmd}{\textsuperscript}
\newcommand{\scr}[1]{\scrcmd{#1}}

\colorcheck{darkspringgreen}
\colorx{red}

\title{Noise or Signal: The Role of Image Backgrounds in Object Recognition}

\author{Kai Xiao \and Logan Engstrom \and Andrew Ilyas \and Aleksander M\k{a}dry\\
  MIT\\
  \texttt{\{kaix, engstrom, ailyas, madry\}@mit.edu} \\
}

\date{}

\begin{document}

\maketitle

\begin{abstract}
We assess the tendency of state-of-the-art object recognition models to
depend on signals from image backgrounds. We create a toolkit for
disentangling foreground and background signal on ImageNet images, and find
that (a) models can achieve non-trivial accuracy by relying on the background
alone, (b) models often misclassify images even in the presence of correctly
classified foregrounds---up to 87.5\% of the time with adversarially chosen
backgrounds, and (c) more accurate models tend to depend on backgrounds less.
Our analysis of backgrounds brings us closer to understanding which
correlations machine learning models use, and how they determine models' out
of distribution performance.
\end{abstract}

\section{Introduction}

Object recognition models are typically trained to minimize loss on a
given dataset, and evaluated by the accuracy they attain on the corresponding test set. In this paradigm,
model performance can be improved by incorporating any generalizing correlation
between images and their labels into decision-making.  
However, the actual model reliability and robustness depend on the specific
set of correlations that is used, and on how those correlations are combined. 
Indeed, outside of the training distribution, model predictions can deviate wildly from
human expectations either due to relying on correlations that humans do not 
perceive~\citep{jetley2018friends,ilyas2019adversarial,jacobsen2019excessive},
or due to overusing
correlations, such as texture~\citep{geirhos2019imagenettrained,baker2018deep} and color~\citep{yip2002contribution}, that humans do use (but to a lesser degree). Characterizing 
the correlations that models depend on thus has important implications for
understanding model behavior, in general.

Image backgrounds are a natural source of correlation between images and
their labels in object recognition.
Indeed, prior work has shown that models may use backgrounds in
classification~\citep{zhang2007local, ribeiro2016why, zhu2017object, rosenfeld2018the, 
barbu2019objectnet, shetty2019not, sagawa2019distributionally}, and suggests
that even human vision makes use of image context for scene and object
recognition~\citep{torralba2003contextual}. In this work, we aim to obtain a deeper understanding
of how current state-of-the-art image classifiers utilize image backgrounds.
Specifically, we investigate the extent to which models rely on them, the implications of this reliance, and how models' use of
backgrounds has evolved over time. Concretely:
\begin{itemize}
    \item We 
        create a variety of datasets that help disentangle the impacts of foreground and
        background signals on classification.
        The test datasets and a public challenge related to them are available at \url{https://github.com/MadryLab/backgrounds_challenge}.
    \item Using the aforementioned toolkit, we characterize models'
        reliance on image backgrounds. We find that image backgrounds alone
        suffice for fairly successful classification
        and that changing background signals decreases average-case performance.
        In fact, we further show that by choosing backgrounds in an adversarial manner, we
        can make standard models misclassify 87.5\% of images
        as the background class.
      \item  We demonstrate that standard models not only use but \emph{require} backgrounds
      for correctly classifying large portions of test sets (35\% on our
      benchmark).
    \item We study the impact of backgrounds on classification for a variety of
      classifiers, and find that more accurate models tend to simultaneously
      exploit background correlations more and have greater robustness
      to changes in image background.
\end{itemize}

\section{Methodology}
To properly gauge image backgrounds' role in image classification, we construct
a synthetic dataset for disentangling background from foreground signal:
\dsname{}.

\paragraph{Base dataset: \dsname{}.} 
We organize a subset of ImageNet into a new dataset with nine coarse-grained classes and
call it \dsname{} (\dsn)~\footnote{These classes are \texttt{dog}, \texttt{bird}, \texttt{vehicle}, \texttt{reptile}, \texttt{carnivore}, \texttt{insect},
\texttt{instrument}, \texttt{primate}, and \texttt{fish}.}.
To create it, we group together ImageNet classes sharing an ancestor in the
WordNet~\citep{miller1995wordnet} hierarchy. We separate out foregrounds and
backgrounds via the annotated bounding boxes provided in ImageNet, and remove
all candidate images whose bounding boxes are unavailable.
We use coarse-grained classes because there are not enough images with bounding boxes
to use the standard labels, and the resulting \dsn dataset has 5045 training and 450 testing
images per class. We describe the dataset creation process in detail in Appendix
\ref{appendix:dataset}. 

\begin{figure}
    \centering
    \includegraphics[width=0.9\linewidth]{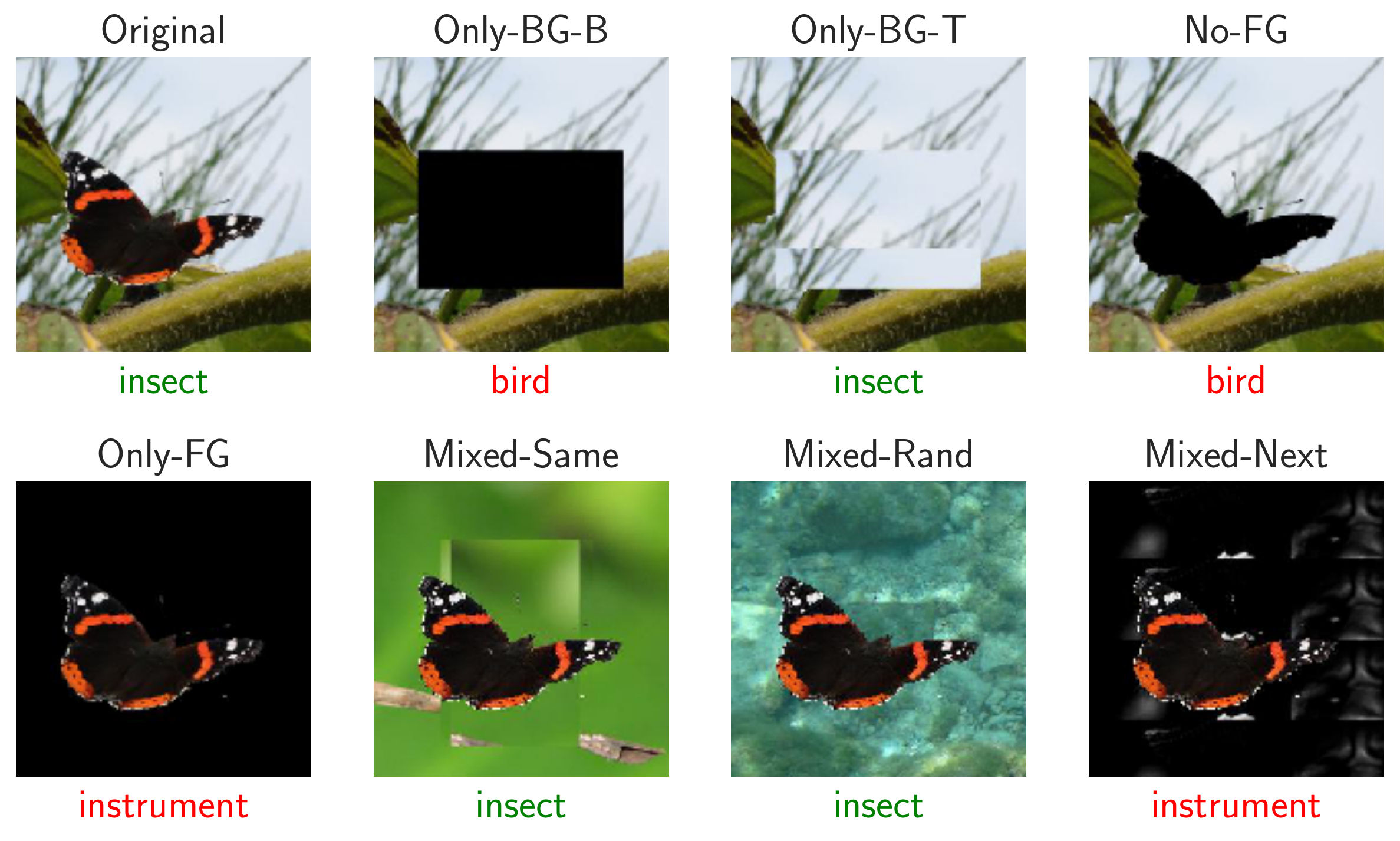}
    \caption{Variations of the synthetic dataset \dsname{}, as described in
      Table~\ref{table:8datasets}. We label each image with its pre-trained
      ResNet-50 classification---green, if corresponding with the original label;
      red, if not. The model correctly classifies the image as ``insect'' when
      given: the original image, only the background, and two cases
      where the original foreground is present but the background changes. Note that, in particular, the model
      fails in two cases when the original foreground is present but the background changes (as in \mnds or \ofgds).}
    \label{fig:butterfly_example}
\end{figure}

\paragraph{Variations of \dsname{}} 
From this base set of images, which we call the \origds version of \dsn, we create seven other synthetic variations designed to understand the impact
of backgrounds. We visualize these variations in
Figure~\ref{fig:butterfly_example}, and provide a detailed reference in Table
\ref{table:8datasets}. These subdatasets of \dsn differ only in how they process
the foregrounds and backgrounds of each constituent image.

\paragraph{Larger dataset: \oads} We finally create a dataset called \oads that consists of all the images in ImageNet corresponding to the classes in \origds (rather than just the images that have associated bounding boxes). We leverage this larger dataset to train better generalizing models.

\begin{table}[ht!]
\centering
\caption{The 8 modified subdatasets created from \dsname{}. The foreground
  detection method refers to how the pixels corresponding to the foreground are
  found. ImageNet annotation refers to the annotated bounding boxes found in
  ImageNet. GrabCut refers to the GrabCut algorithm~\citep{rother2004grabcut} as
  implemented in OpenCV2. Random backgrounds in the last three datasets are
  taken from \obgtds. For more details see Appendix~\ref{appendix:dataset}.}
\label{table:8datasets}
\resizebox{\linewidth}{!}{
\begin{tabular}{@{}llll@{}}
\toprule
\textbf{Name} & \textbf{Foreground} & \textbf{Background} & \textbf{Foreground Detection Method}\\
\midrule
\origds & Unmodified & Unmodified & --- \\
\obgbds & Black & Unmodified & ImageNet Annotation \\
\obgtds & Tiled background & Unmodified & ImageNet Annotation \\
\nfgds & Black & Unmodified & GrabCut \\
\ofgds & Unmodified & Black & GrabCut \\
\msds & Unmodified & Random background of the same class & GrabCut\\
\mrds & Unmodified & Random background of a random class & GrabCut\\
\mnds & Unmodified & Random background of the next class & GrabCut\\
\bottomrule
\end{tabular}
}
\end{table}

\label{sec:methodology}

\section{Quantifying Reliance on Background Signals} 
\label{section:reliance}
With \dsname{} in hand, we now assess the role of image backgrounds in
classification.

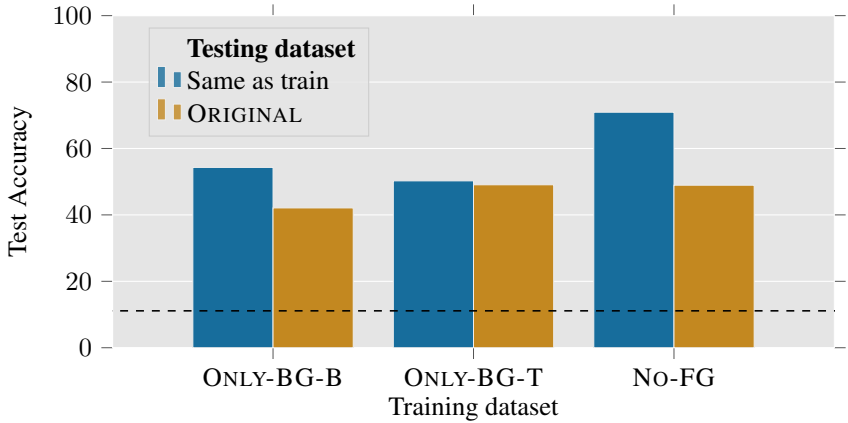
\begin{figure}
    \centering
    \begin{tikzpicture}

\definecolor{color0}{rgb}{0.0906862745098039,0.425980392156863,0.611274509803922}
\definecolor{color1}{rgb}{0.76421568627451,0.531862745098039,0.125980392156863}

\begin{axis}[
axis background/.style={fill=white!89.8039215686275!black},
height={6cm},
width={0.8\textwidth},
axis line style={white},
legend cell align={left},
legend style={fill opacity=0.8, draw opacity=1, text opacity=1, 
at={(0.05,0.8)}, anchor=west, draw=white!80!black, 
fill=white!89.8039215686275!black},
tick align=outside,
x grid style={white},
enlarge x limits=0.1,
xlabel={Training dataset},
xmajorticks=true,
xmin=-0.5, xmax=2.5,
xtick style={color=white!33.3333333333333!black},
xtick={0,1,2},
xticklabels={\textsc{Only-BG-B},\textsc{Only-BG-T},\textsc{No-FG}},
y grid style={white},
ylabel={Test Accuracy},
ymajorgrids,
ymajorticks=true,
ymin=0, ymax=100,
ytick style={color=white!33.3333333333333!black}
]
\addlegendimage{empty legend}
\addlegendentry{{\bf Testing dataset}}

\draw[draw=white!93.3333333333333!black,fill=color0,very thin] (axis cs:-0.4,0) rectangle (axis cs:0,54.2962951660156);
\addlegendimage{ybar,ybar legend,draw=white!93.3333333333333!black,fill=color0,very thin};
\addlegendentry{Same as train}

\draw[draw=white!93.3333333333333!black,fill=color0,very thin] (axis cs:0.6,0) rectangle (axis cs:1,50.2469139099121);
\draw[draw=white!93.3333333333333!black,fill=color0,very thin] (axis cs:1.6,0) rectangle (axis cs:2,70.9135818481445);
\draw[draw=white!93.3333333333333!black,fill=color1,very thin] (axis cs:-2.77555756156289e-17,0) rectangle (axis cs:0.4,42.0987663269043);
\addlegendimage{ybar,ybar legend,draw=white!93.3333333333333!black,fill=color1,very thin};
\addlegendentry{\textsc{Original}}

\draw[draw=white!93.3333333333333!black,fill=color1,very thin] (axis cs:1,0) rectangle (axis cs:1.4,49.0617294311523);
\draw[draw=white!93.3333333333333!black,fill=color1,very thin] (axis cs:2,0) rectangle (axis cs:2.4,48.9382705688477);
\addplot [line width=1.08pt, white!26!black, forget plot]
table {%
-0.2 nan
-0.2 nan
};
\addplot [line width=1.08pt, white!26!black, forget plot]
table {%
0.8 nan
0.8 nan
};
\addplot [line width=1.08pt, white!26!black, forget plot]
table {%
1.8 nan
1.8 nan
};
\addplot [line width=1.08pt, white!26!black, forget plot]
table {%
0.2 nan
0.2 nan
};
\addplot [line width=1.08pt, white!26!black, forget plot]
table {%
1.2 nan
1.2 nan
};
\addplot [line width=1.08pt, white!26!black, forget plot]
table {%
2.2 nan
2.2 nan
};
\addplot [semithick, black, dashed, forget plot]
table {%
-1.0 11.1111111111111
3.0 11.1111111111111
};
\end{axis}

\end{tikzpicture}
    \caption{We train models on each of the ``background-only'' datasets, then
      evaluate each on its corresponding test set as well as the \origds test
      set. Even though the model only learns from background signal, it achieves
      (much) better than random performance on \textit{both} the corresponding
      test set and \origds. Here, random guessing would give 11.11\% (the dotted
      line).}
    \label{fig:sec3_1_barplot}
\end{figure}

\paragraph{Backgrounds suffice for classification.} Prior work has found that
models are able to make accurate predictions based on backgrounds alone; we begin by directly quantifying this ability. 
Looking at the \obgtds, \obgbds, and \nfgds datasets, we find (cf.
Figure~\ref{fig:sec3_1_barplot}) that models trained on these background-only
training sets generalize reasonably well to
both their corresponding test sets and the \origds test set (around 40-50\% for every model, far
above the baseline of 11\% representing random classification).
Our results confirm that image backgrounds contain signal that models can
accurately classify with.

\renewcommand{\rothead}[2][60]{\makebox[9mm][c]{\rotatebox{#1}{\makecell[c]{#2}}}}%
\definecolor{Gray}{gray}{0.8}
\newcolumntype{a}{>{\columncolor{Gray}}c}
\begin{table}
\centering
\caption{Performance of state-of-the-art computer vision models on select
test sets of \dsname{}. We include both pre-trained ImageNet models and
models of different architectures that we train on \oads. The \bggap
is defined as the difference in test accuracy between \msds and \mrds and
helps assess the tendency of such models to rely on background signal.
Architectures are sorted by their test accuracies on ImageNet and \origds
for pre-trained and \oads-trained models, respectively. Shaded in grey are the two
architectures that can be directly compared across datasets (ResNet-50 and
Wide-ResNet-50x2).}  
\label{table:pretrained_models}
\resizebox{\textwidth}{!}{
\begin{tabular}{@{}lccaac c ccaac@{}}
  \toprule
&  \multicolumn{5}{c}{{\bf Pre-trained on ImageNet}} & & \multicolumn{5}{c}{{\bf Trained on \oads}} \\
\cmidrule(lr){2-6} \cmidrule(lr){8-12}
Test dataset & \rothead{MobileNet-v3} & \rothead{EfficientNet-b0} & 
  \rothead{ResNet-50} & \rothead{WRN-50x2} & \rothead{DPN-92} & &
  \rothead{AlexNet} & \rothead{ShuffleNet} & \rothead{ResNet-50} &
  \rothead{WRN-50x2} & \rothead{VGG16-BN} \\ \toprule
ImageNet & 67.9\% & 77.2\% & 77.6\% & 78.5\% & 80.0\% & & \multicolumn{5}{c}{------} \\
\origds  & 91.0\% & 95.6\% & 96.2\% & 95.8\% & 96.8\% & 
         & 86.7\% & 95.7\% & 96.3\% & 97.2\% & 97.6\% \\ 
\oads    & 90.0\% & 94.3\% & 95.0\% & 95.5\% & 96.0\% & 
         & 83.1\% & 93.2\% & 94.6\% & 95.2\% & 96.0\% \\ \midrule
\obgtds  & 15.7\% & 11.9\% & 17.8\% & 20.7\% & 20.6\% & 
         & 41.5\% & 43.6\% & 43.6\% & 45.1\% & 45.7\% \\ \midrule
\msds    & 69.7\% & 79.7\% & 82.3\% & 81.7\% & 85.4\% & 
         & 76.2\% & 86.7\% & 89.9\% & 90.6\% & 91.0\% \\ 
\mrds    & 56.1\% & 67.8\% & 76.3\% & 73.0\% & 77.6\% & 
         & 54.2\% & 69.4\% & 75.6\% & 78.0\% & 78.0\% \\ \midrule
BG-gap   & 13.6\% & 11.9\% &  6.0\% &  8.7\% &  7.8\% & 
         & 22.0\% & 17.3\% & 14.3\% & 12.6\% & 13.0\% \\ 
\bottomrule
\end{tabular}
}
\end{table}

\paragraph{Models exploit background signal for classification.} We
discover that models can misclassify due to background signal, especially
when the background class does not match that of the foreground.  
As a demonstration, we study model accuracies on the \mrds dataset, where image
backgrounds are randomized and thus provide no information about the correct label. By comparing test accuracies on \mrds and \msds~\footnote{
\msds controls for artifacts from image processing present in \mrds. For further discussion, see 
Appendix~\ref{appendix:results}.},
where images have class-consistent backgrounds,
we can measure classifiers' dependence on the correct background.
We denote the resulting accuracy gap between \msds and \mrds as the \bggap;
this difference represents the drop in model accuracy due to changing the
class signal from the background.
In Table~\ref{table:pretrained_models}, we observe a \bggap of
13-22\% and 6-14\% for models trained on \oads and ImageNet, respectively, %
suggesting that backgrounds often mislead state-of-the-art models \textit{even
  when the correct foreground is present}.

Our results indicate that ImageNet-trained models are less dependent on
backgrounds than their \oads-trained counterparts---they have a smaller (but
still significant) \bggap, and perform worse when predicting solely
based on backgrounds (i.e., on the \obgtds dataset). An initial hypothesis could
be that ImageNet-trained models' lesser dependence results from having either
(a) a more fine-grained class structure, or (b) more datapoints than \oads.
However, a preliminary investigation (Appendix~\ref{appendix:explaining_imagenet_bggap}) ablating both
fine-grainedness and dataset size does not find evidence supporting either explanation.
Therefore, understanding why pre-trained ImageNet models rely less on
backgrounds than \oads models remains an open question.

\begin{figure}
    \centering
    \includegraphics[width=\linewidth]{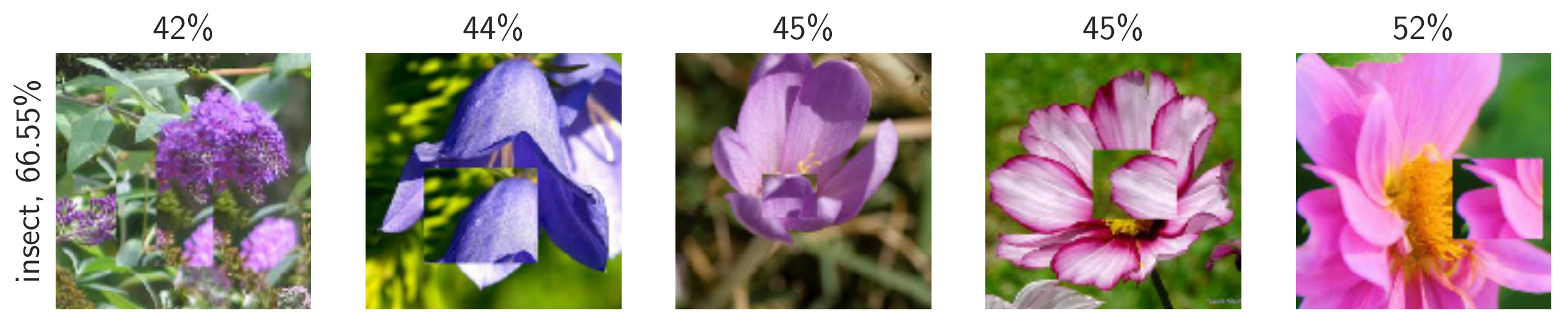}
    \caption{The adversarial backgrounds that most frequently fool \oads-trained models into classifying a
      given foreground as insect, ordered by the percentage of foregrounds
      fooled. The total portion of images that can be fooled (by any background
      from this class) is 66.55\%.}
    \label{fig:most_fooling}
\end{figure}

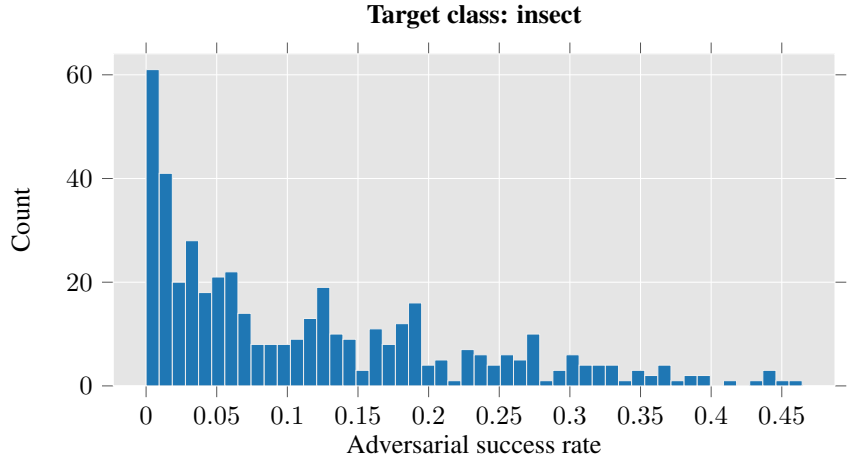
\begin{figure}
    \centering
    \begin{tikzpicture}

\definecolor{color0}{rgb}{0.12156862745098,0.466666666666667,0.705882352941177}

\begin{axis}[
height={6cm},
width={0.8\textwidth},
xticklabel style={
        /pgf/number format/fixed,
        /pgf/number format/precision=5
},
scaled x ticks={false},
axis background/.style={fill=white!89.8039215686275!black},
axis line style={white},
tick align=outside,
title={{\bf Target class: insect}},
x grid style={white},
xlabel={Adversarial success rate},
xmajorgrids,
xmajorticks=true,
xmin=-0.0232222222222222, xmax=0.487666666666667,
xtick style={color=white!33.3333333333333!black},
y grid style={white},
ylabel={Count},
ymajorgrids,
ymajorticks=true,
ymin=0, ymax=64.05,
ytick style={color=white!33.3333333333333!black}
]
\draw[draw=white!93.3333333333333!black,fill=color0,very thin] (axis cs:0,0) rectangle (axis cs:0.00928888888888889,61);
\draw[draw=white!93.3333333333333!black,fill=color0,very thin] (axis cs:0.00928888888888889,0) rectangle (axis cs:0.0185777777777778,41);
\draw[draw=white!93.3333333333333!black,fill=color0,very thin] (axis cs:0.0185777777777778,0) rectangle (axis cs:0.0278666666666667,20);
\draw[draw=white!93.3333333333333!black,fill=color0,very thin] (axis cs:0.0278666666666667,0) rectangle (axis cs:0.0371555555555556,28);
\draw[draw=white!93.3333333333333!black,fill=color0,very thin] (axis cs:0.0371555555555556,0) rectangle (axis cs:0.0464444444444444,18);
\draw[draw=white!93.3333333333333!black,fill=color0,very thin] (axis cs:0.0464444444444444,0) rectangle (axis cs:0.0557333333333333,21);
\draw[draw=white!93.3333333333333!black,fill=color0,very thin] (axis cs:0.0557333333333333,0) rectangle (axis cs:0.0650222222222222,22);
\draw[draw=white!93.3333333333333!black,fill=color0,very thin] (axis cs:0.0650222222222222,0) rectangle (axis cs:0.0743111111111111,14);
\draw[draw=white!93.3333333333333!black,fill=color0,very thin] (axis cs:0.0743111111111111,0) rectangle (axis cs:0.0836,8);
\draw[draw=white!93.3333333333333!black,fill=color0,very thin] (axis cs:0.0836,0) rectangle (axis cs:0.0928888888888889,8);
\draw[draw=white!93.3333333333333!black,fill=color0,very thin] (axis cs:0.0928888888888889,0) rectangle (axis cs:0.102177777777778,8);
\draw[draw=white!93.3333333333333!black,fill=color0,very thin] (axis cs:0.102177777777778,0) rectangle (axis cs:0.111466666666667,9);
\draw[draw=white!93.3333333333333!black,fill=color0,very thin] (axis cs:0.111466666666667,0) rectangle (axis cs:0.120755555555556,13);
\draw[draw=white!93.3333333333333!black,fill=color0,very thin] (axis cs:0.120755555555556,0) rectangle (axis cs:0.130044444444444,19);
\draw[draw=white!93.3333333333333!black,fill=color0,very thin] (axis cs:0.130044444444444,0) rectangle (axis cs:0.139333333333333,10);
\draw[draw=white!93.3333333333333!black,fill=color0,very thin] (axis cs:0.139333333333333,0) rectangle (axis cs:0.148622222222222,9);
\draw[draw=white!93.3333333333333!black,fill=color0,very thin] (axis cs:0.148622222222222,0) rectangle (axis cs:0.157911111111111,3);
\draw[draw=white!93.3333333333333!black,fill=color0,very thin] (axis cs:0.157911111111111,0) rectangle (axis cs:0.1672,11);
\draw[draw=white!93.3333333333333!black,fill=color0,very thin] (axis cs:0.1672,0) rectangle (axis cs:0.176488888888889,8);
\draw[draw=white!93.3333333333333!black,fill=color0,very thin] (axis cs:0.176488888888889,0) rectangle (axis cs:0.185777777777778,12);
\draw[draw=white!93.3333333333333!black,fill=color0,very thin] (axis cs:0.185777777777778,0) rectangle (axis cs:0.195066666666667,16);
\draw[draw=white!93.3333333333333!black,fill=color0,very thin] (axis cs:0.195066666666667,0) rectangle (axis cs:0.204355555555556,4);
\draw[draw=white!93.3333333333333!black,fill=color0,very thin] (axis cs:0.204355555555556,0) rectangle (axis cs:0.213644444444444,5);
\draw[draw=white!93.3333333333333!black,fill=color0,very thin] (axis cs:0.213644444444444,0) rectangle (axis cs:0.222933333333333,1);
\draw[draw=white!93.3333333333333!black,fill=color0,very thin] (axis cs:0.222933333333333,0) rectangle (axis cs:0.232222222222222,7);
\draw[draw=white!93.3333333333333!black,fill=color0,very thin] (axis cs:0.232222222222222,0) rectangle (axis cs:0.241511111111111,6);
\draw[draw=white!93.3333333333333!black,fill=color0,very thin] (axis cs:0.241511111111111,0) rectangle (axis cs:0.2508,4);
\draw[draw=white!93.3333333333333!black,fill=color0,very thin] (axis cs:0.2508,0) rectangle (axis cs:0.260088888888889,6);
\draw[draw=white!93.3333333333333!black,fill=color0,very thin] (axis cs:0.260088888888889,0) rectangle (axis cs:0.269377777777778,5);
\draw[draw=white!93.3333333333333!black,fill=color0,very thin] (axis cs:0.269377777777778,0) rectangle (axis cs:0.278666666666667,10);
\draw[draw=white!93.3333333333333!black,fill=color0,very thin] (axis cs:0.278666666666667,0) rectangle (axis cs:0.287955555555556,1);
\draw[draw=white!93.3333333333333!black,fill=color0,very thin] (axis cs:0.287955555555556,0) rectangle (axis cs:0.297244444444444,3);
\draw[draw=white!93.3333333333333!black,fill=color0,very thin] (axis cs:0.297244444444444,0) rectangle (axis cs:0.306533333333333,6);
\draw[draw=white!93.3333333333333!black,fill=color0,very thin] (axis cs:0.306533333333333,0) rectangle (axis cs:0.315822222222222,4);
\draw[draw=white!93.3333333333333!black,fill=color0,very thin] (axis cs:0.315822222222222,0) rectangle (axis cs:0.325111111111111,4);
\draw[draw=white!93.3333333333333!black,fill=color0,very thin] (axis cs:0.325111111111111,0) rectangle (axis cs:0.3344,4);
\draw[draw=white!93.3333333333333!black,fill=color0,very thin] (axis cs:0.3344,0) rectangle (axis cs:0.343688888888889,1);
\draw[draw=white!93.3333333333333!black,fill=color0,very thin] (axis cs:0.343688888888889,0) rectangle (axis cs:0.352977777777778,3);
\draw[draw=white!93.3333333333333!black,fill=color0,very thin] (axis cs:0.352977777777778,0) rectangle (axis cs:0.362266666666667,2);
\draw[draw=white!93.3333333333333!black,fill=color0,very thin] (axis cs:0.362266666666667,0) rectangle (axis cs:0.371555555555556,4);
\draw[draw=white!93.3333333333333!black,fill=color0,very thin] (axis cs:0.371555555555556,0) rectangle (axis cs:0.380844444444444,1);
\draw[draw=white!93.3333333333333!black,fill=color0,very thin] (axis cs:0.380844444444444,0) rectangle (axis cs:0.390133333333333,2);
\draw[draw=white!93.3333333333333!black,fill=color0,very thin] (axis cs:0.390133333333333,0) rectangle (axis cs:0.399422222222222,2);
\draw[draw=white!93.3333333333333!black,fill=color0,very thin] (axis cs:0.399422222222222,0) rectangle (axis cs:0.408711111111111,0);
\draw[draw=white!93.3333333333333!black,fill=color0,very thin] (axis cs:0.408711111111111,0) rectangle (axis cs:0.418,1);
\draw[draw=white!93.3333333333333!black,fill=color0,very thin] (axis cs:0.418,0) rectangle (axis cs:0.427288888888889,0);
\draw[draw=white!93.3333333333333!black,fill=color0,very thin] (axis cs:0.427288888888889,0) rectangle (axis cs:0.436577777777778,1);
\draw[draw=white!93.3333333333333!black,fill=color0,very thin] (axis cs:0.436577777777778,0) rectangle (axis cs:0.445866666666667,3);
\draw[draw=white!93.3333333333333!black,fill=color0,very thin] (axis cs:0.445866666666667,0) rectangle (axis cs:0.455155555555556,1);
\draw[draw=white!93.3333333333333!black,fill=color0,very thin] (axis cs:0.455155555555556,0) rectangle (axis cs:0.464444444444444,1);
\end{axis}

\end{tikzpicture}
    \caption{Histogram of insect backgrounds grouped by how often they cause
      (non-insect) foregrounds to be classified as insect by a \oads-trained
      model. We visualize the five backgrounds that fool the classifier on the
      largest percentage of images in Figure~\ref{fig:most_fooling}.}
    \label{fig:adv_hist}
\end{figure}

\paragraph{Models are vulnerable to adversarial backgrounds.} To
understand how worst-case backgrounds impact models' performance, we evaluate
model robustness to adversarially chosen backgrounds. We find that 87.5\% of
foregrounds are susceptible to such backgrounds; that is, for
these foregrounds, there is a background that causes the classifier to classify
the resulting foreground-background combination as the background class. For a
finer grained look, we also evaluate image backgrounds based on their attack
success rate (ASR), i.e., how frequently they cause models to predict the
(background) class in the presence of a conflicting foreground class. As an
example, Figure~\ref{fig:most_fooling} shows the five
backgrounds with the highest ASR for the insect class---these backgrounds (extracted from
insect images in \origds) fool a \oads-trained ResNet-50 model into predicting
insect on up to 52\% of non-insect foregrounds. We plot a histogram of ASR
over all insect backgrounds in Figure~\ref{fig:adv_hist}---it has a long tail.
Similar results are observed for other classes as well (cf. Appendix
\ref{appendix:adv}).

\paragraph{Training on \mrds reduces background dependence.}
Next, we explore how to reduce models' dependence on background. To this end, we train
models on \mrds, a synthetic dataset where background signals are decorrelated
from class labels. As we would expect, \mrds-trained models extract less signal
from backgrounds: evaluation results show that \mrds
models perform poorly (15\% accuracy---barely higher than random) on datasets with only
backgrounds and no foregrounds (\obgtds or \obgbds).

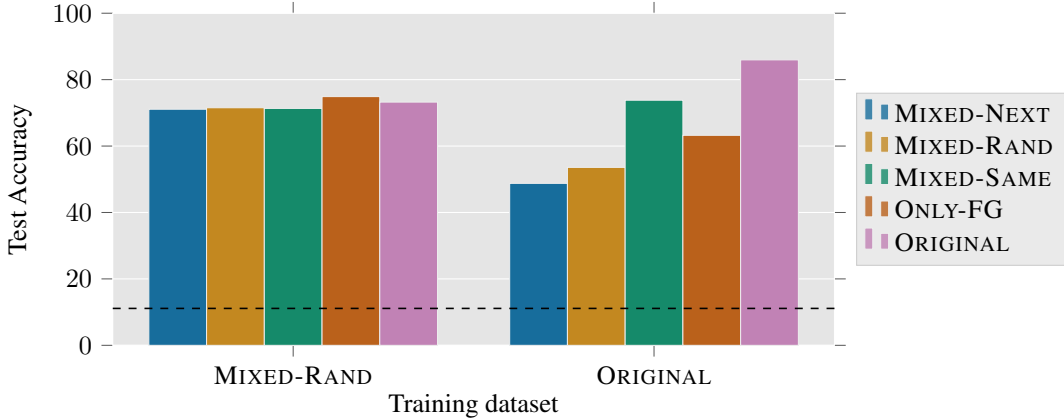
\begin{figure}
    \centering
    \begin{tikzpicture}

\definecolor{color0}{rgb}{0.0906862745098039,0.425980392156863,0.611274509803922}
\definecolor{color1}{rgb}{0.76421568627451,0.531862745098039,0.125980392156863}
\definecolor{color2}{rgb}{0.084313725490196,0.543137254901961,0.416666666666667}
\definecolor{color3}{rgb}{0.730882352941177,0.380882352941177,0.104411764705882}
\definecolor{color4}{rgb}{0.758823529411765,0.511764705882353,0.711764705882353}

\begin{axis}[
height={6cm},
width={0.8\textwidth},
axis background/.style={fill=white!89.8039215686275!black},
axis line style={white},
legend cell align={left},
legend style={fill opacity=0.8, draw opacity=1, text opacity=1, 
at={(1.03,0.5)}, anchor=west, draw=white!80!black, 
fill=white!89.8039215686275!black},
tick align=outside,
x grid style={white},
xlabel={Training dataset},
xmajorticks=true,
xmin=-0.5, xmax=1.5,
xtick style={color=white!33.3333333333333!black},
xtick={0,1},
xticklabels={\textsc{Mixed-Rand},\textsc{Original}},
y grid style={white},
ylabel={Test Accuracy},
ymajorgrids,
ymajorticks=true,
ymin=0, ymax=100,
ytick style={color=white!33.3333333333333!black}
]
\draw[draw=white!93.3333333333333!black,fill=color0,very thin] (axis cs:-0.4,0) rectangle (axis cs:-0.24,71.0864181518555);
\addlegendimage{ybar,ybar legend,draw=white!93.3333333333333!black,fill=color0,very thin};
\addlegendentry{\textsc{Mixed-Next}}

\draw[draw=white!93.3333333333333!black,fill=color0,very thin] (axis cs:0.6,0) rectangle (axis cs:0.76,48.7654304504395);
\draw[draw=white!93.3333333333333!black,fill=color1,very thin] (axis cs:-0.24,0) rectangle (axis cs:-0.08,71.5308609008789);
\addlegendimage{ybar,ybar legend,draw=white!93.3333333333333!black,fill=color1,very thin};
\addlegendentry{\textsc{Mixed-Rand}}

\draw[draw=white!93.3333333333333!black,fill=color1,very thin] (axis cs:0.76,0) rectangle (axis cs:0.92,53.5802459716797);
\draw[draw=white!93.3333333333333!black,fill=color2,very thin] (axis cs:-0.08,0) rectangle (axis cs:0.08,71.3333358764648);
\addlegendimage{ybar,ybar legend,draw=white!93.3333333333333!black,fill=color2,very thin};
\addlegendentry{\textsc{Mixed-Same}}

\draw[draw=white!93.3333333333333!black,fill=color2,very thin] (axis cs:0.92,0) rectangle (axis cs:1.08,73.8024673461914);
\draw[draw=white!93.3333333333333!black,fill=color3,very thin] (axis cs:0.08,0) rectangle (axis cs:0.24,74.8888854980469);
\addlegendimage{ybar,ybar legend,draw=white!93.3333333333333!black,fill=color3,very thin};
\addlegendentry{\textsc{Only-FG}}

\draw[draw=white!93.3333333333333!black,fill=color3,very thin] (axis cs:1.08,0) rectangle (axis cs:1.24,63.2345695495605);
\draw[draw=white!93.3333333333333!black,fill=color4,very thin] (axis cs:0.24,0) rectangle (axis cs:0.4,73.2345657348633);
\addlegendimage{ybar,ybar legend,draw=white!93.3333333333333!black,fill=color4,very thin};
\addlegendentry{\textsc{Original}}

\draw[draw=white!93.3333333333333!black,fill=color4,very thin] (axis cs:1.24,0) rectangle (axis cs:1.4,85.9506149291992);
\addplot [line width=1.08pt, white!26!black, forget plot]
table {%
-0.32 nan
-0.32 nan
};
\addplot [line width=1.08pt, white!26!black, forget plot]
table {%
0.68 nan
0.68 nan
};
\addplot [line width=1.08pt, white!26!black, forget plot]
table {%
-0.16 nan
-0.16 nan
};
\addplot [line width=1.08pt, white!26!black, forget plot]
table {%
0.84 nan
0.84 nan
};
\addplot [line width=1.08pt, white!26!black, forget plot]
table {%
0 nan
0 nan
};
\addplot [line width=1.08pt, white!26!black, forget plot]
table {%
1 nan
1 nan
};
\addplot [line width=1.08pt, white!26!black, forget plot]
table {%
0.16 nan
0.16 nan
};
\addplot [line width=1.08pt, white!26!black, forget plot]
table {%
1.16 nan
1.16 nan
};
\addplot [line width=1.08pt, white!26!black, forget plot]
table {%
0.32 nan
0.32 nan
};
\addplot [line width=1.08pt, white!26!black, forget plot]
table {%
1.32 nan
1.32 nan
};
\addplot [semithick, black, dashed, forget plot]
table {%
-0.5 11.1111111111111
1.5 11.1111111111111
};
\end{axis}

\end{tikzpicture}
    \caption{We compare the test performance of a model trained on the
    synthetic \mrds dataset with a model trained on \origds. We evaluate
    these models on variants of \dsn{} that contain identical
    foregrounds. For the \origds-trained model, test performance decreases
    significantly when the background signal is modified during testing. However, the \mrds-trained model is robust to background
    changes, albeit at the cost of lower accuracy on images from
    \origds.}
    \label{fig:mixedrand_vs_orig_comparison}
\end{figure}

Indeed, such models are also more accurate on datasets where backgrounds do not
match foregrounds. In Figure~\ref{fig:mixedrand_vs_orig_comparison}, we observe
that a \mrds-trained model has 17.3\% higher accuracy than its \origds-trained
counterpart on \mrds, and 22.3\% higher accuracy on \mnds, a dataset where
background signals class-consistently mismatch foregrounds. (Recall that \mnds
images have foregrounds from class $y$ mixed with backgrounds from class $y+1$,
labeled as class $y$.) The \mrds-trained model also has little variation (at
most 3.8\%) in accuracy across all five test sets that contain the correct
foreground.

\begin{figure}
    \centering
    \includegraphics[width=0.95\linewidth]{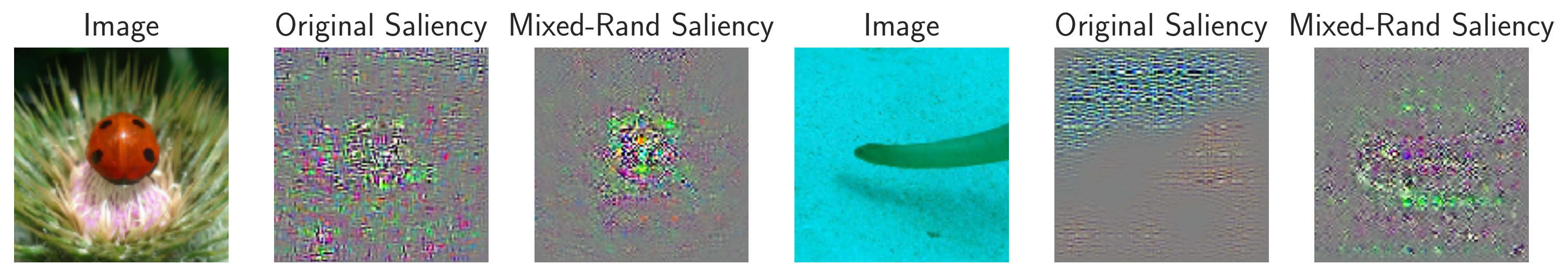}
    \caption{Saliency maps for the the \origds and \mrds models on two
      images. As expected, the \mrds model appears to place more importance on foreground pixels.}
    \label{fig:saliency}
\end{figure}

Qualitatively, the \mrds-trained model also appears to place more relative
importance on foreground pixels than the \origds-trained model; the saliency
maps of the two models in Figure~\ref{fig:saliency} show that the \mrds-trained
model's saliency maps highlight more foreground pixels than those of
\origds-trained models.

\paragraph{A fine grained look at dependence on backgrounds.}
We now analyze models' reliance on backgrounds at an image-by-image level and ask: for which
images does introducing backgrounds help or hurt classifiers' performance? To this end, for each image in \origds, we decompose how models use
foreground and background signals by examining classifiers' predictions on the
corresponding image in \mrds and \obgtds. Here, we use the \mrds and \obgtds
predictions as a proxy for which class the foreground and background signals
(alone) point towards, respectively. We categorize each image based on how its
background and foreground signals impact classification; we list the categories
in Table~\ref{table:fine_grained_defs} and show the counts for each category as
a histogram per classifier in Figure~\ref{fig:apx_fine_grained_bars}. Our
results show that while few backgrounds induce misclassification (see Appendix~\ref{appendix:misleading_backgrounds} for examples), a large
fraction of images require backgrounds for correct
classification---approximately 35\% on the \origds trained classifiers, as
calculated by combining the ``BG Required'' and ``BG+FG Required'' categories.

\begin{table}[ht!]
\renewcommand{\arraystretch}{1.2}
\centering
\caption{Prediction categories we study for a given image-model pair. For a
  given image, a model can make differing predictions based on the presence or
  absence of its foreground/background. We label each possible case based on how
  the background classification relates to the original image classification and
  the foreground classification. To proxy classifying full images, foregrounds,
  and backgrounds separately, we classify on \origds, \mrds, and \obgtds
  (respectively). ``BG Irrelevant'' demarcates images where the foreground
  classification result is identical to that of the full image (in terms of
  correctness).}
\label{table:fine_grained_defs}

\begin{tabular}{@{}lllll@{}}
\toprule
Label      & Correct Prediction & Correct Prediction & Correct Prediction \\
{}   & on Full Image  & on Foreground & on Background  \\
\midrule
BG Required            & \darkspringgreencheck & \redx   & \darkspringgreencheck \\
BG Fools            & \redx  & \darkspringgreencheck  & \redx    \\
BG+FG Required       & \darkspringgreencheck & \redx   & \redx    \\
BG+FG Fools          & \redx  & \darkspringgreencheck  & \darkspringgreencheck    \\
BG Irrelevant          & \darkspringgreencheck/\redx  & \darkspringgreencheck/\redx  &  --- \\
\bottomrule
\end{tabular}
\end{table}

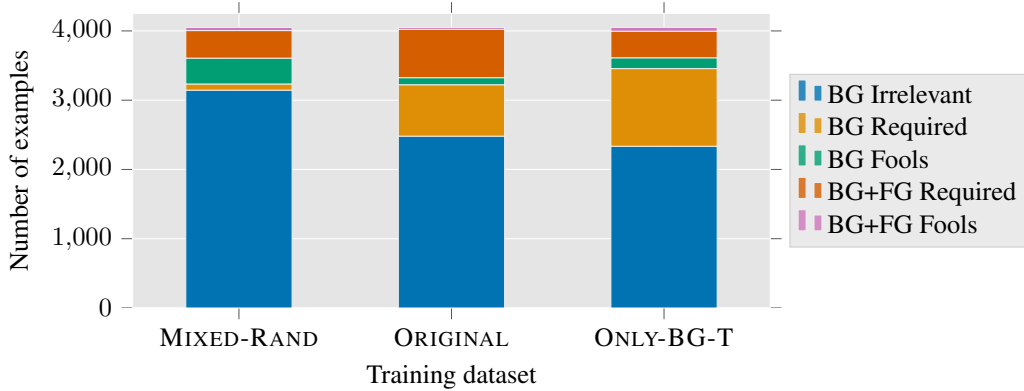
\begin{figure}[h]
    \centering
    \pgfplotsset{compat=1.3}
\begin{tikzpicture}

\definecolor{color0}{rgb}{0.00392156862745098,0.450980392156863,0.698039215686274}
\definecolor{color1}{rgb}{0.870588235294118,0.56078431372549,0.0196078431372549}
\definecolor{color2}{rgb}{0.00784313725490196,0.619607843137255,0.450980392156863}
\definecolor{color3}{rgb}{0.835294117647059,0.368627450980392,0}
\definecolor{color4}{rgb}{0.8,0.470588235294118,0.737254901960784}

\begin{axis}[
height={5.5cm},
width={0.72\textwidth},
axis background/.style={fill=white!89.8039215686275!black},
axis line style={white},
legend cell align={left},
legend style={fill opacity=0.8, draw opacity=1, text opacity=1, at={(1.03,0.5)}, anchor=west, draw=white!80!black, fill=white!89.8039215686275!black},
tick align=outside,
x grid style={white},
xlabel={Training dataset},
xmajorgrids,
xmajorticks=true,
xmin=-0.5, xmax=2.5,
xtick style={color=white!33.3333333333333!black},
xtick={0,1,2},
xticklabels={\textsc{Mixed-Rand},\textsc{Original},\textsc{Only-BG-T}},
y grid style={white},
ylabel={Number of examples},
ymajorgrids,
ymajorticks=true,
ymin=0, ymax=4252.5,
ytick style={color=white!33.3333333333333!black}
]
\draw[draw=white!93.3333333333333!black,fill=color0,very thin] (axis cs:-0.25,0) rectangle (axis cs:0.25,3145);
\addlegendimage{ybar,ybar legend,draw=white!93.3333333333333!black,fill=color0,very thin};
\addlegendentry{BG Irrelevant}

\draw[draw=white!93.3333333333333!black,fill=color0,very thin] (axis cs:0.75,0) rectangle (axis cs:1.25,2481);
\draw[draw=white!93.3333333333333!black,fill=color0,very thin] (axis cs:1.75,0) rectangle (axis cs:2.25,2335);
\draw[draw=white!93.3333333333333!black,fill=color1,very thin] (axis cs:-0.25,3145) rectangle (axis cs:0.25,3234);
\addlegendimage{ybar,ybar legend,draw=white!93.3333333333333!black,fill=color1,very thin};
\addlegendentry{BG Required}

\draw[draw=white!93.3333333333333!black,fill=color1,very thin] (axis cs:0.75,2481) rectangle (axis cs:1.25,3223);
\draw[draw=white!93.3333333333333!black,fill=color1,very thin] (axis cs:1.75,2335) rectangle (axis cs:2.25,3456);
\draw[draw=white!93.3333333333333!black,fill=color2,very thin] (axis cs:-0.25,3234) rectangle (axis cs:0.25,3607);
\addlegendimage{ybar,ybar legend,draw=white!93.3333333333333!black,fill=color2,very thin};
\addlegendentry{BG Fools}

\draw[draw=white!93.3333333333333!black,fill=color2,very thin] (axis cs:0.75,3223) rectangle (axis cs:1.25,3325);
\draw[draw=white!93.3333333333333!black,fill=color2,very thin] (axis cs:1.75,3456) rectangle (axis cs:2.25,3612);
\draw[draw=white!93.3333333333333!black,fill=color3,very thin] (axis cs:-0.25,3607) rectangle (axis cs:0.25,4005);
\addlegendimage{ybar,ybar legend,draw=white!93.3333333333333!black,fill=color3,very thin};
\addlegendentry{BG+FG Required}

\draw[draw=white!93.3333333333333!black,fill=color3,very thin] (axis cs:0.75,3325) rectangle (axis cs:1.25,4023);
\draw[draw=white!93.3333333333333!black,fill=color3,very thin] (axis cs:1.75,3612) rectangle (axis cs:2.25,3996);
\draw[draw=white!93.3333333333333!black,fill=color4,very thin] (axis cs:-0.25,4005) rectangle (axis cs:0.25,4050);
\addlegendimage{ybar,ybar legend,draw=white!93.3333333333333!black,fill=color4,very thin};
\addlegendentry{BG+FG Fools}

\draw[draw=white!93.3333333333333!black,fill=color4,very thin] (axis cs:0.75,4023) rectangle (axis cs:1.25,4050);
\draw[draw=white!93.3333333333333!black,fill=color4,very thin] (axis cs:1.75,3996) rectangle (axis cs:2.25,4050);
\end{axis}

\end{tikzpicture}
    \caption{
    We categorize each test set image based on how a model classifies the full image, the background alone, and the foreground alone (cf. Table~\ref{table:fine_grained_defs}). The model trained on \origds needs the background for correct classification on 35\% of images (measured by adding ``BG Required'' and ``BG+FG Required), while a model trained on \mrds is much less reliant on background. The model trained on \obgtds requires the background most, as expected; however, the model often misclassifies both the full image and the background, so the ``BG Irrelevant'' subset is still sizable.
    }
    \label{fig:apx_fine_grained_bars}
\end{figure}

\paragraph{Further insights derived from \dsn{} are discussed in the Appendix~\ref{appendix:results}.} We focus on key findings in this section, but also include more comprehensive results and examples of other questions that can be explored by using the toolkit of \dsn{} in the Appendix.

\section{Benchmark Progress and Background Dependence}
\label{section:adaptive_overfitting}
In the previous sections, we demonstrated that standard image classification models
exploit signals from backgrounds. Considering that these models result from
progress on standard computer vision benchmarks, a natural
question is: {\em to what extent have improvements on image classification
  benchmarks resulted from exploiting background correlations?} 
  And relatedly, {\em how has model robustness to misleading background signals evolved over time?}

\begin{figure}
    \centering
    \input{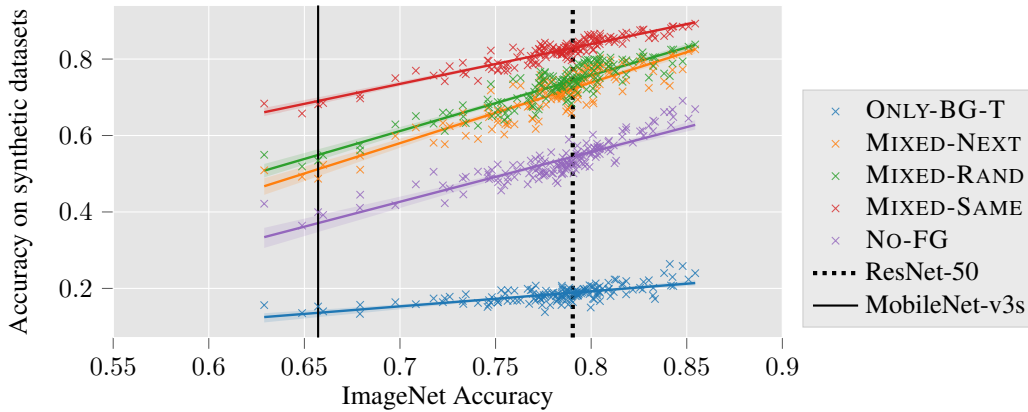}
    \caption{Measuring progress on each of the synthetic \dsname{} datasets with
      respect to progress on the standard ImageNet test set. Higher accuracy on
      ImageNet generally corresponds to higher accuracy on each of the
      constructed datasets, but the rate at which accuracy grows varies based on
      the types of features present in each dataset.
      Each pre-trained model corresponds to a vertical line on the plot---we
      mark ResNet-50 and MobileNet-v3s models for reference.}
    \label{fig:all_models}
\end{figure}

As a first step towards answering these questions, we study the progress made by ImageNet models on our synthetic
\dsn{} dataset variations. 
In Figure~\ref{fig:all_models} we plot accuracy on our synthetic datasets against ImageNet accuracy for each of the architectures considered. 
As evidenced by the lines of best fit in
Figure~\ref{fig:all_models}, accuracy increases on the original ImageNet
benchmark generally correspond to accuracy increases on all of the synthetic datasets.
This includes the \obgds datasets---indicating that models {\em do} improve at
extracting correlations from image backgrounds.

Indeed, the \obgds trend observed in Figure~\ref{fig:all_models} suggests that either (a) image classification 
models can only attain their reported accuracies in the presence of background signals;
or (b) these models carry an implicit bias
towards features in the background, as a result of optimization technique, model
class, etc.---in this case, we may need explicit regularization (e.g., through
distributionally robust optimization~\cite{sagawa2019distributionally} or
related techniques) to obtain models invariant to these background features.

Still, models' {\em relative} improvement in accuracy across dataset variants is promising---models improve on classifying \obgtds at a
slower (absolute) rate than \mrds, \msds and \mnds. Furthermore, the
performance gap between the \mixedds datasets and the others (most notably,
between \mrds and \msds; between \mnds and \mrds; and consequently between
\mnds and \msds) trends towards closing, indicating that models not only are 
becoming better at using foreground features, but also are becoming more
robust to misleading background features (\mrds and \mnds).

Overall, the accuracy trends observed from testing ImageNet models on our
synthetic datasets reveal that better models (a) are capable of exploiting 
background correlations, but (b) are increasingly robust to changes in
background, suggesting that invariance to background features may not necessarily come at
the cost of benchmark accuracy.

\section{Related Work}
\label{sec:related_work}
We focus our discussion here on the works most closely related to ours,
specifically those investigating contextual bias or background dependence in
computer vision (for a more extensive survey of and explicit comparison
to prior work, see Appendix~\ref{appendix:rel_work_long}). Prior work has explored
the more general phenomenon of contextual
bias~\citep{torralba2011unbiased,khosla2012undoing,choi2012context,shetty2019not},
including studying its prevalence and developing methods to mitigate it.
For image backgrounds specifically, prior works show that background correlations can
be predictive~\citep{torralba2003contextual}, and that backgrounds can influence
model decisions to varying degrees~\citep{zhang2007local, ribeiro2016why,
zhu2017object, beery2018recognition, rosenfeld2018the, barbu2019objectnet, sagawa2019distributionally}. 
The work most similar to ours is that of~\citet{zhu2017object}, who also analyze
ImageNet classification (focusing on the older, AlexNet model). They find that
the AlexNet model achieve small but non-trivial test accuracy on a dataset similar
to our \obgbds dataset. While sufficient for establishing that backgrounds can
be used for classification, this dataset also introduces biases by adding large
black rectangular patches to all of the images. In comparison
to~\citep{zhu2017object} (and the prior works mentioned earlier): 
(a) we create a more extensive toolkit that allows us to
measure not just model performance without foregrounds but also the relative
influence of foregrounds and backgrounds on model predictions; (b) we control for
the effect of image artifacts via the \msds dataset; (c) we study model
robustness to adversarial backgrounds; (d) we study a larger and more recent set
of image classifiers~\citep{he2016deep,zagoruyko2016wide,tan2019efficientnet}, and how improvements they give on ImageNet relate to background correlations.

\section{Discussion and Conclusion}
In this work, we study the extent to which classifiers rely on image
backgrounds. To this end, we create a toolkit for measuring the precise role of background
and foreground signal that involves constructing new test datasets that contain different
amounts of each. Through these datasets we establish both the usefulness of
background signal and the tendency of our models to depend on backgrounds, even
when relevant foreground features are present. Our results show that our models
are not robust to changes in the background, either in the adversarial case, or in the
average case.

As most ImageNet images have human-recognizable foreground objects, our models appear to rely on background more than
humans on that dataset. The fact that models can be fooled by adversarial background
changes on 87.5\% of all images highlights how poorly
computer vision models may perform in an out-of-distribution setting. However,
contextual information like the background can still be useful in certain settings.
After all, humans do use backgrounds as context in visual processing, and the
background may be necessary if the foreground is blurry or distorted
\citep{torralba2003contextual}. Therefore, reliance on background is a nuanced question that merits
further study.

On one hand, our findings provide evidence that models succeed by using
background correlations, which may be undesirable in some applications. On the
other hand, we find that advances in classifiers have given rise to models that use
foregrounds more effectively and are more robust to changes in the background.
To obtain even more robust models, we may want to draw inspiration from successes in
training on the \mrds dataset (a dataset designed to neutralize background
signal---cf. Table~\ref{table:8datasets}), related data-augmentation techniques \citep{shetty2019not},
and training algorithms like distributionally robust optimization \citep{sagawa2019distributionally} and model-based robust learning \cite{robey2020modelbased}. Overall, the
toolkit and findings in this work help us to better understand models and
to monitor our progress toward the goal of reliable machine learning.

\section*{Acknowledgements}

Thanks to Kuo-An ``Andy'' Wei, John Cherian, and anonymous conference reviewers for helpful comments on earlier versions of this work. The authors would also like to thank Guillaume LeClerc, Sam Park, and Kuo-An Wei for help with data labeling. KX was supported by the NDSEG Fellowship. LE was supported by the NSF Graduate Research Fellowship. AI was supported by the  Open Phil AI Fellowship. Work supported in part by the NSF grants CCF-1553428, CNS-1815221, and the Microsoft Corporation. This material is based upon work supported by the Defense Advanced Research Projects Agency (DARPA) under Contract No. HR001120C0015.

\printbibliography

\clearpage
\appendix

\section{Datasets Details}
\label{appendix:dataset}

We choose the following 9 high-level classes.
\begin{table}[ht]
\centering
\begin{tabular}{l l r}
\midrule
\textbf{Class} & WordNet ID & Number of \\
{} & {} & sub-classes \\
\midrule
\textbf{Dog} & n02084071 & 116 \\
\textbf{Bird} & n01503061 &  52 \\
\textbf{Vehicle} & n04576211 &  42 \\
\textbf{Reptile} & n01661091 &  36 \\
\textbf{Carnivore} & n02075296 &  35 \\
\textbf{Insect} & n02159955 &  27 \\
\textbf{Instrument} & n03800933 &  26 \\
\textbf{Primate} & n02469914 &  20 \\
\textbf{Fish} & n02512053 &  16 \\
\midrule
\end{tabular}
\caption{The 9 classes of \dsname{}.}
\label{table:9classes}
\end{table}

\begin{figure}
    \centering
    \includegraphics[width=0.75\linewidth]{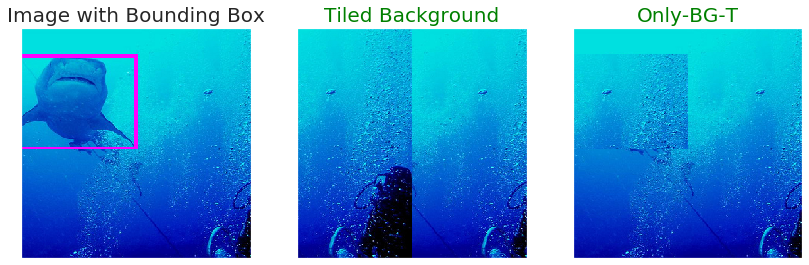}
    \caption{Visualization of how \obgtds is created.}
    \label{fig:tiled_explanation}
\end{figure}

All datasets used in the paper are balanced by randomly removing images
from classes that are over-represented. We only keep as many images as the
smallest post-modification synthetic dataset, so all synthetic datasets
(except \oads) have the same number of images. We also use a custom GUI to
manually process the test set to improve data quality. For \oads, the only difference from using the corresponding classes in the original ImageNet dataset is that we balance the dataset. \\ \\
\textbf{For all images}: we apply the following filters before adding each image to our datasets.
\begin{itemize}
    \item The image must have bounding box annotations.
    \item For simplicity, each image must have exactly one bounding box. A large majority of images that have bounding box annotations satisfy this.
\end{itemize}
\textbf{For images needing a properly segmented foreground}: This includes the 3 \mixedds datasets, \ofgds, and \nfgds. We filter out images based on the following criteria.
\begin{itemize}
    \item Because images are cropped before they are fed into models, we require that less than 50\% of the bounding box is removed by the crop, to ensure that the foreground still exists. Almost all images pass this filter.
    \item The OpenCV function \texttt{cv2.grabCut} (used to extract the foreground shape) must work on the image. We remove images where it fails.
    \item For the test set only, we manually remove images with foreground segmentations that retain a significant portion of the background signal.
    \item For the test set only, we manually remove foreground segmentations that are very bad (e.g. the segmentation selects part of the image, and that part doesn't contain the foreground object).
\end{itemize}
\textbf{For images needing only background signal}: This includes \obgbds and \obgtds. In this case, we apply the following criteria:
\begin{itemize}
    \item The bounding box must not be too big (more than 90\% of the image). The intent here is to avoid \obgbds images being just a large black rectangle.
    \item For the test set only, we manually remove \obgds images that still have an instance of the class even after removing the bounding box. This occurs when the bounding boxes are imperfect or incomplete (e.g. only one of two dogs in an image is labeled with a bounding box).
\end{itemize}
\textbf{Creating the \obgtds dataset}: We first make a ``tiled'' version of the background by finding the largest rectangular strip (horizontal or vertical) outside the bounding box, and tiling the entire image with that strip. We then replace the removed foreground with the tiled background. A visual example is provided in Figure~\ref{fig:tiled_explanation}. We purposefully choose not to use deep-learning-based inpainting techniques such as \citep{shetty2018adversarial} to replace the removed foreground, as such methods could lead to biases that the inpainting model has learned from the data. For example, an inpainting model may learn that the best way to inpaint a missing chunk of a flower is to place an insect there, which is something we want to avoid.

\section{Explaining the Decreased \bggap of pre-trained ImageNet models}
\label{appendix:explaining_imagenet_bggap}

We investigate two possible explanations for why pre-trained ImageNet models have a smaller \bggap than models trained on \dsname{}. Understanding this phenomenon can help inform how models should be trained to be more background-robust. We find slight improvements to background-robustness from training on more fine-grained classes, and even smaller improvements from training on larger datasets. These two factors thus do not completely explain the increased background-robustness of pre-trained ImageNet models---understanding this further is an interesting empirical mystery to study.

\subsection{The Effect of Fine-grainedness on the \bggap}
\label{appendix:fine_grainedness}
One possible explanation is that training models to distinguish between finer-grained classes forces them to focus more on the foreground, which contains relevant features for making those fine-grained distinctions, than the background, which may be fairly similar across sub-classes of a high-level class. This suggests that asking models to solve more fine-grained tasks could improve model robustness to background changes.

To test the effect of fine-grainedness on \dsname{}, we make a related dataset called \dsnamebal{} that uses the same 9 high-level classes and can be cleanly modified into more fine-grained versions. Specifically, for \dsnamebal{} we choose exactly 16 sub-classes for each high-level class, for a total of 144 ImageNet classes. To create successively more fine-grained versions of the \dsnamebal{} dataset, we group every $n$ sub-classes together into a higher-level class, for $n \in \{1,2,4,8,16\}$. Here, $n=1$ corresponds to keeping all 144 ImageNet classes as they are, while $n=16$ corresponds to only having 9 high-level classes, like \dsname{}. Because we keep all images from those original ImageNet classes, this dataset is the same size as \oads{}.

\begin{figure}[h]
    \centering
    \includegraphics[width=0.9\linewidth]{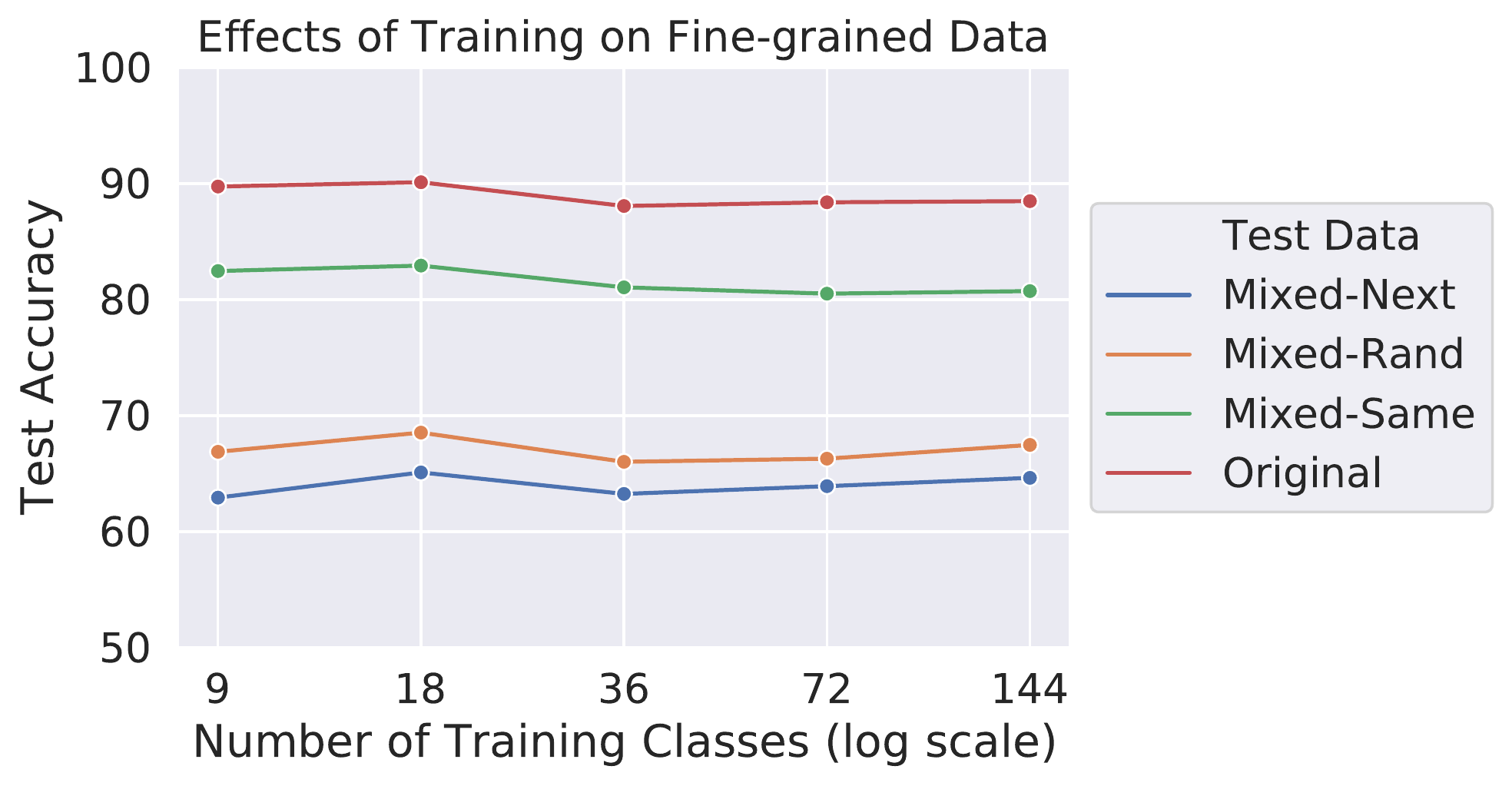}
    \caption{We train models on \dsnamebal{} at different levels of fine-grainedness (more training classes is more fine-grained). The \bggap, or the difference between the test accuracies on \msds and \mrds, decreases as we make the classification task more fine-grained, but the decrease is small compared to the size of the \bggap.}
    \label{fig:fine_grainedness_plot}
\end{figure}

We train models on \dsnamebal{} at different levels of fine-grainedness and evaluate the \bggap of those models in Figure~\ref{fig:fine_grainedness_plot}. We find that fine-grained models have a smaller \bggap as well as better performance on \mnds, but the improvement is very slight and also comes at the cost of decreased accuracy on \origds. The \bggap of the most fine-grained classifier is 2.3\% smaller than the \bggap of the most coarse-grained classifier, showing that fine-grainedness does improve background-robustness. However, the improvement is still small compared to the size of the \bggap (which is 13.3\% for the fine-grained classifier).

\subsection{The Effect of Larger Dataset Size on the \bggap}
\label{appendix:more_data}
\begin{figure}[h]
    \centering
    \includegraphics[width=0.9\linewidth]{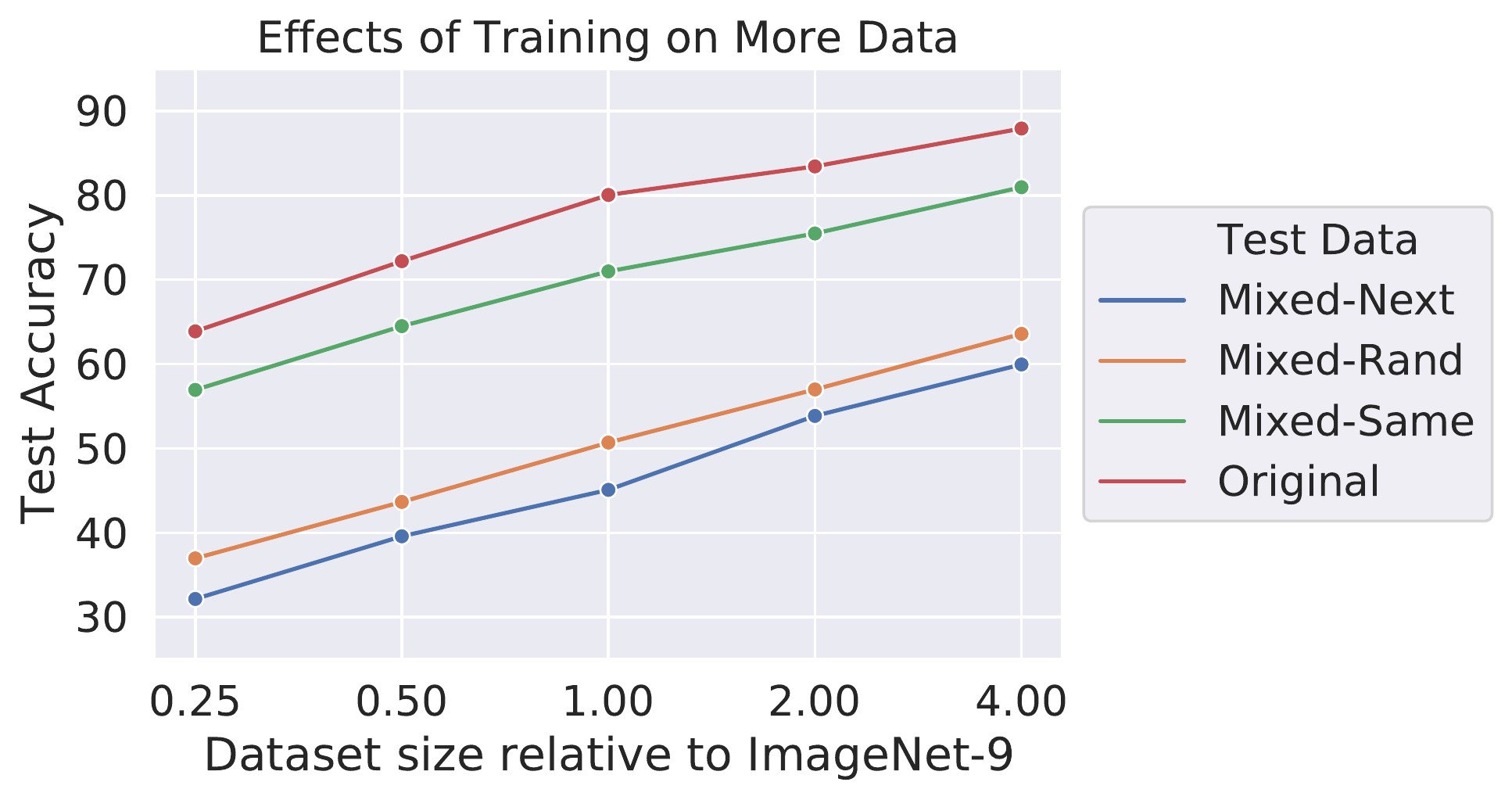}
    \caption{We train models on different-sized subsets of \dsnamebal{}. The largest training set we use is the full \dsnamebal{} dataset, which is 4 times larger than \dsname{}. While performance on all test datasets improves as the amount of training data increases, the \bggap has almost the same size regardless of the amount of training data used.}
    \label{fig:more_data_plot}
\end{figure}

A second possible explanation for why pre-trained ImageNet models have a smaller \bggap is that training on larger datasets is important for background-robustness. To evaluate this possibility, we train models on different-sized subsets of \dsnamebal{}. The largest dataset we train on is the full \dsnamebal{} dataset, which is 4 times as large as \dsn{}, and the smallest is 1/4 as large as \dsn{}. Figure~\ref{fig:more_data_plot} shows that increasing the dataset size does increase overall performance but does not noticeably decrease the \bggap.

It is possible that even more data for these classes (more than is available from ImageNet) is needed to improve background-robustness, or that more training data (from other classes) is the cause of the increased background-robustness of pre-trained ImageNet models. Understanding this further would be an interesting direction.

\subsection{Summary of methods investigated to reduce the \bggap}
\begin{figure}
    \centering
    \includegraphics[width=0.75\linewidth]{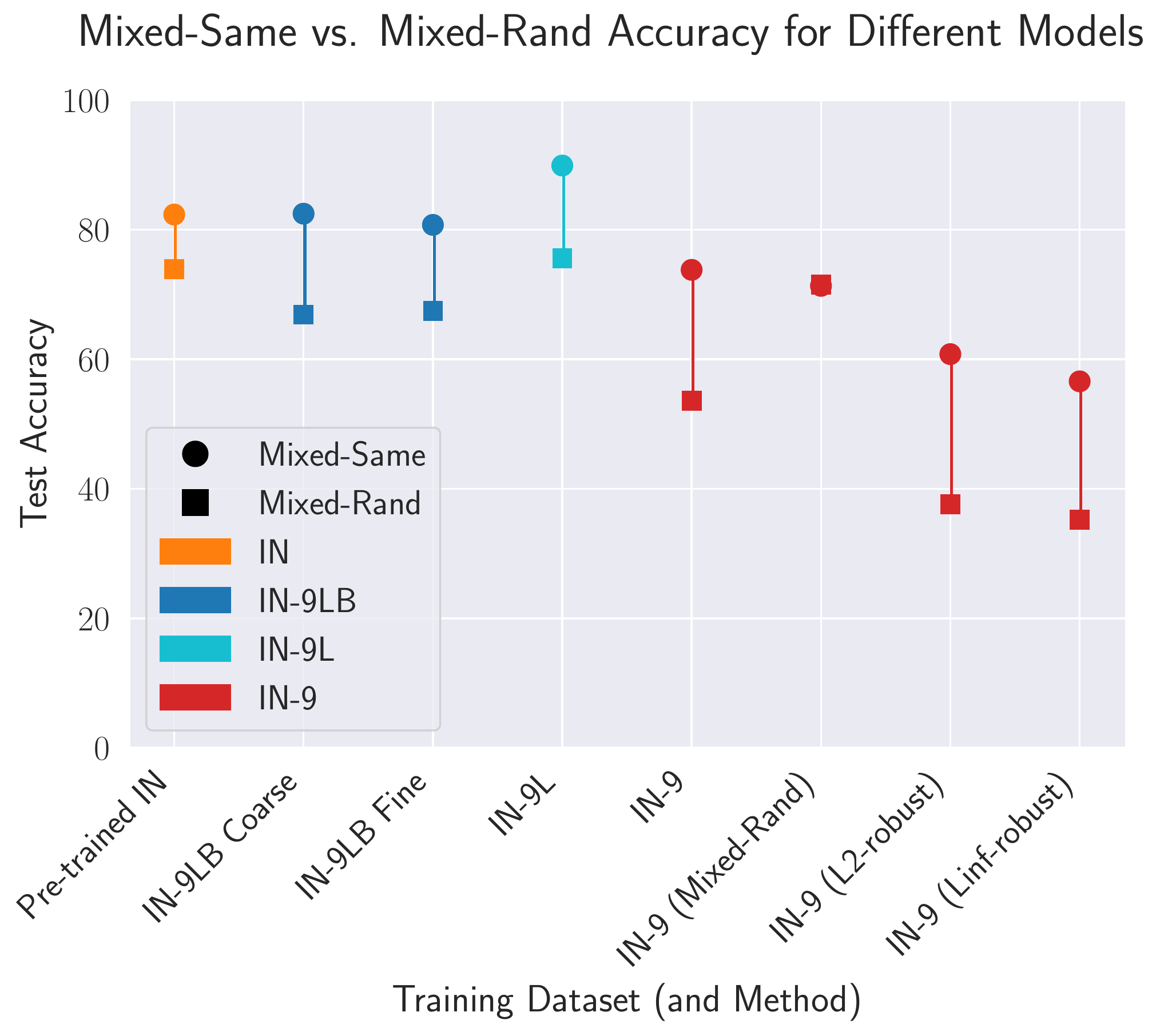}
    \caption{We compare various different methods of training models and measure their \bggap, or the difference between \msds and \mrds test accuracy. We find that (1) Pre-trained IN models have surprisingly small \bggap. (2) Increasing fine-grainedness (\dsnamebal Coarse vs. \dsnamebal Fine) and dataset size (\dsn vs. \oads) decreases the \bggap only slightly. (3) $\ell_p$-robust training does not help. (4) Training on \mrds (cf. Section~\ref{section:reliance} appears to be the most effective strategy for reducing the \bggap. For such a model, the \msds and \mrds accuracies are nearly identical.}
    \label{fig:mixed_same_vs_mixed_rand}
\end{figure}

In Figure~\ref{fig:mixed_same_vs_mixed_rand}, we compare the \bggap of ResNet-50 models trained on different datasets and with different methods to a ResNet-50 pre-trained on ImageNet. We explore $\ell_p$-robust training, increasing dataset size, and making the classification task more fine-grained, and find that none of these methods reduces the \bggap as much as pre-training on ImageNet. The only method that reduces the \bggap significantly more is training on \mrds. Furthermore, the same trends hold true if we measure the difference between \msds and \mnds as opposed to the \bggap (the difference between \msds and \mrds).

\section{Training details}
For all models, we use fairly standard training settings for ImageNet-style models. We train for $200$ epochs using SGD with a batch size of $256$, a learning rate of $0.1$ (with learning rate drops every $50$ epochs), a momentum parameter of $0.9$, a weight decay of $1$e$-4$, and data augmentation (random resized crop, random horizontal flip, and color jitter). Unless specified, we always use a standard ResNet-50 architecture \citep{he2016deep}. For the experiment depicted in Figure~\ref{fig:more_data_plot}, we found that using a smaller learning rate of $0.01$ was necessary for training to converge on the smaller training sets. Thus, we used that same learning rate for all models in Figure~\ref{fig:more_data_plot}.

\section{Additional Evaluation Results}
\label{appendix:results}

We include full results of training models on every synthetic \dsn variation and then testing them on every synthetic \dsn variation in Table \ref{table:apx_allresults}.
In addition to being more comprehensive, this table can help answer a variety of questions, of which we provide two examples here.

\textbf{How much information is leaked from the size of the foreground bounding box?}

The scale of an object already gives signal correlated with the object class \citep{torralba2003contextual}. Even though they are designed to avoid having foreground signal, the background-only datasets \obgbds and \obgtds may inadvertently leak information about object scale due to the bounding box sizes being recognizable.

To gauge the extent of this leakage, we can measure how models trained on datasets where only the foreground signal has useful correlation (\mrds or \ofgds) perform on the background-only test sets. We find that there is small signal leakage from bounding box size alone---a model trained on \ofgds achieves about 23\% background-only test accuracy, suggesting that it is able to exploit the signal leakage to some degree. A model trained on \mrds achieves about 15\% background-only test accuracy, just slightly better than random, perhaps because it is harder for models to measure (and thus, make use of) object scale when training on \mrds.

The existence of a small amount of information leakage in this case shows the importance of comparing \msds (as opposed to just \origds) with \mrds and \mnds when assessing model dependence on backgrounds. Indeed, the \mixedds datasets may contain (1) image processing artifacts, such as rough edges from the foreground processing, and (2) small traces of the original background. This makes it important to control for both factors when measuring how models react to varying background signal.

\textbf{How does more training data affect model performance with and without object shape?}

We already show closely related results on the effect of more training data on the \bggap in Figure~\ref{fig:more_data_plot}. Here, we compare model test performance on the \nfgds and \obgbds test sets. Both replace the foreground with black, but only \nfgds retains the foreground shape.

By comparing the models trained on \origds and \oads (4x more training data), we find that
\begin{enumerate}
    \item The \origds-trained model performs similarly on \nfgds and \obgbds, indicating that it does not use object shape effectively.
    \item The \oads-trained model performs about 13\% better on \nfgds than \obgbds, showing that it uses object shape more effectively.
\end{enumerate}
Thus, this suggests that more training data may allow models to learn to use object shape more effectively.
Understanding this phenomena further could help inform model training and dataset collection
if the goal is to train models that are able to leverage shape effectively.

\begin{table}[ht!]
\centering
\resizebox{\textwidth}{!}{
\begin{tabular}{l|rrrrrrrrr}
\toprule
\textbf{Trained on} &  {} & {} & {} & {} & \textbf{Test Dataset} & {} & {} & {} & {} \\
{} &  \mnds &  \mrds &  \msds &  \nfgds &  \obgbds &  \obgtds &  \ofgds &  \origds &  \oads \\
\midrule
\mnds   & 78.07 & 53.28 & 48.49 & 16.20 & 11.19 &  8.22 & 59.60 & 52.32 & 46.44 \\
\mrds   & 71.09 & 71.53 & 71.33 & 26.72 & 15.33 & 14.62 & 74.89 & 73.23 & 67.53 \\
\msds   & 45.41 & 51.36 & 74.40 & 39.85 & 35.19 & 41.58 & 61.65 & 75.01 & 69.21 \\
\nfgds        & 13.70 & 18.74 & 42.79 & 70.91 & 36.79 & 42.52 & 31.48 & 48.94 & 47.62 \\
\obgbds    & 10.35 & 15.41 & 38.37 & 37.85 & 54.30 & 42.54 & 21.38 & 42.10 & 41.01 \\
\obgtds    & 11.48 & 17.09 & 45.80 & 40.84 & 38.49 & 50.25 & 19.19 & 49.06 & 47.94 \\
\ofgds      & 33.04 & 35.88 & 47.63 & 27.90 & 23.58 & 22.59 & 84.20 & 54.62 & 51.50 \\
\origds     & 48.77 & 53.58 & 73.80 & 42.22 & 32.94 & 40.54 & 63.23 & 85.95 & 80.38 \\
\oads & 71.21 & 75.60 & 89.90 & 55.78 & 34.02 & 43.60 & 84.12 & 96.32 & 94.61 \\
\bottomrule
\end{tabular}
}
\caption{The test accuracies, in percentages, of models trained on all variants of
\dsname{}.}
\label{table:apx_allresults}
\end{table}

\textbf{What about other ways of modifying the background signal?}

\begin{figure}
    \centering
    \includegraphics[width=0.74\linewidth]{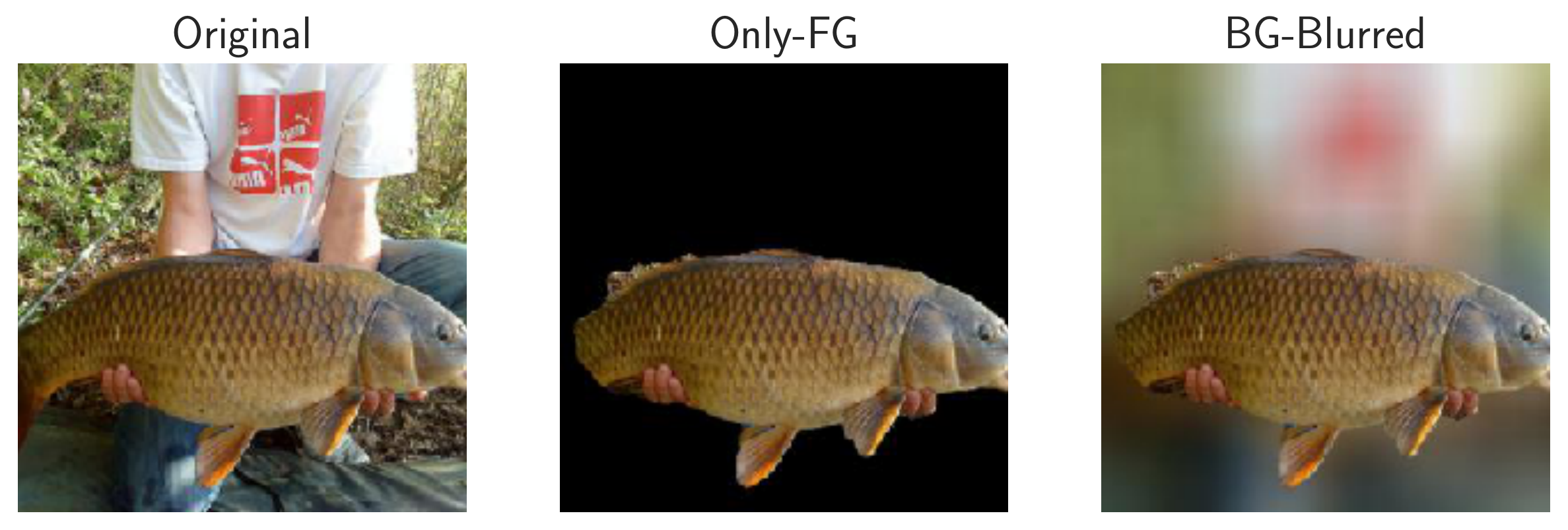}
    \caption{Backgrounds can also be modified in other ways; for example, it can be blurred. Our evaluations on this dataset show similar results.}
    \label{fig:blurred_example}
\end{figure}

One can modify the background in various other ways---for example, instead of replacing the background with black as in \ofgds, the background can be blurred  as in the \textsc{BG-Blurred} image of Figure~\ref{fig:blurred_example}.
As expected, blurred backgrounds are still slightly correlated with the correct class.
Thus, test accuracies for standard models on this dataset are higher than on \ofgds, but lower than on \msds (which has signal from random class-aligned backgrounds that are \emph{not} blurred).
While we do not investigate all possible methods of modifying background signal, we believe that the variations we do examine in \dsname{} already improve our understanding of how background signals matter.
Investigating other variations could provide an even more nuanced understanding of what parts of the background are most important.

\section{Additional Related Works and Explicit Comparisons}
\label{appendix:rel_work_long}

There has been prior work on mitigating contextual bias in image classification, the influence of background signals on various datasets, and techniques like foreground segmentation that we leverage.

\textbf{Mitigating Contextual Bias:}
\citep{khosla2012undoing} focuses on mitigating dataset-specific contextual bias and proposes learning SVMs with both general weights and dataset-specific weights, while \citep{choi2012context} creates an out-of-context detection task with 209 out-of-context images and suggests using graphical models to solve it. \citep{shetty2019not} focuses on the role of co-occurring objects as context in the MS-COCO dataset, and uses object removal to show that (a) models can still predict a removed object when only co-occurring objects are shown, and (b) special data-augmentation can mitigate this.

\textbf{Understanding the influence of backgrounds:}
For contextual bias from image backgrounds specifically, prior works have observed that the background of an image can influence model decisions to varying degrees.
In particular, \citep{zhang2007local} find that (a) a bag-of-features object detection algorithm depends on image backgrounds in the PASCAL dataset and (b) using this algorithm on a training set with varying backgrounds leads to better generalization.
\citep{beery2018recognition, barbu2019objectnet} collect new test datasets of animals and objects, respectively. \citep{barbu2019objectnet} focus on object classes that also exist in ImageNet, and their new test set contains objects photographed in front of unconventional backgrounds and in unfamiliar orientations. Both works show that computer vision models experience significant accuracy drops when trained on data with one set of backgrounds and tested on data with another.
\citep{sagawa2019distributionally} create a small synthetic dataset of Waterbirds, where waterbirds and landbirds from one dataset are combined with water and land backgrounds from another. They show that a model's reliance on spurious correlations with the background can be harmful for small subgroups of data where those spurious correlations no longer hold (e.g. landbirds on water backgrounds). Furthermore, they propose using distributionally robust optimization to reduce reliance on spurious correlations with the background, but their method assumes that the spurious correlation can be precisely specified in advance. 
\citep{rosenfeld2018the} analyzes background dependence for object detection (as opposed to classification) models on the MS-COCO dataset. They transplant an object from one image to another image, and find that object detection models may detect the transplanted object differently depending on its location, and that the transplanted object may also cause mispredictions on other objects in the image.

\textbf{Explicit Comparison to Prior Works}: In comparison to prior works, our work contributes the following.
\begin{itemize}
\item We develop a toolkit for analyzing the background dependence of ImageNet classifiers, the most common benchmark for computer vision progress. Only \citep{zhu2017object}, which we compare to in Section~\ref{sec:related_work}, also focuses on ImageNet.
\item The test datasets we create separate and mix foreground and background signals in various ways (cf. Table~\ref{table:8datasets}), allowing us to study the sensitivity of models to these signals in a more fine-grained manner.
\item Our toolkit for separating foreground and background can be applied without human-annotated foreground segmentation, which prior works on MS-COCO and Waterbirds rely on. This is important because foreground segmentation annotations are hard to collect and do not exist for ImageNet.
\item We study the extent of background dependence in the extreme case of adversarial backgrounds.
\item We focus on better vision models, including ResNet \citep{he2016deep}, Wide ResNet\citep{zagoruyko2016wide}, and EfficientNet \citep{tan2019efficientnet}.
\item We evaluate how improvements on the ImageNet benchmark have affected background dependence (cf. Section \ref{section:adaptive_overfitting}).
\end{itemize}

\textbf{Foreground Segmentation and Image Inpainting:} In order to create \dsn and its variants, we rely on OpenCV's implementation of the foreground segmentation algorithm GrabCut \citep{rother2004grabcut}. Foreground segmentation is a branch of computer vision that seeks to automatically extract the foreground from an image \citep{harville2001foreground}. After finding the foreground, we remove it and simply replace the foreground with copies of parts of the background. Other works solve this problem, called image inpainting, either using exemplar-based methods \citep{criminisi2004region} or using deep learning \cite{yu2018generative, shetty2018adversarial}. \citep{shetty2018adversarial} both detects the foreground for removal and inpaints the removed region. However, more advanced inpaintings techniques can be slow and inaccurate when the region that must be inpainted is relatively large \citep{shetty2018adversarial}, which is the case for many ImageNet images. Exploring better ways of segementing the foreground and inpainting the removed foreground could improve our analysis toolkit further.

\clearpage

\section{Additional examples of synthetic datasets}

We randomly sample an image from each class, and display all
synthetic variations of that image, as well as the predictions of a
pre-trained ResNet-50 (trained on \oads) on each variant.

\begin{figure}[h]
\centering
\includegraphics[width=0.9\linewidth]{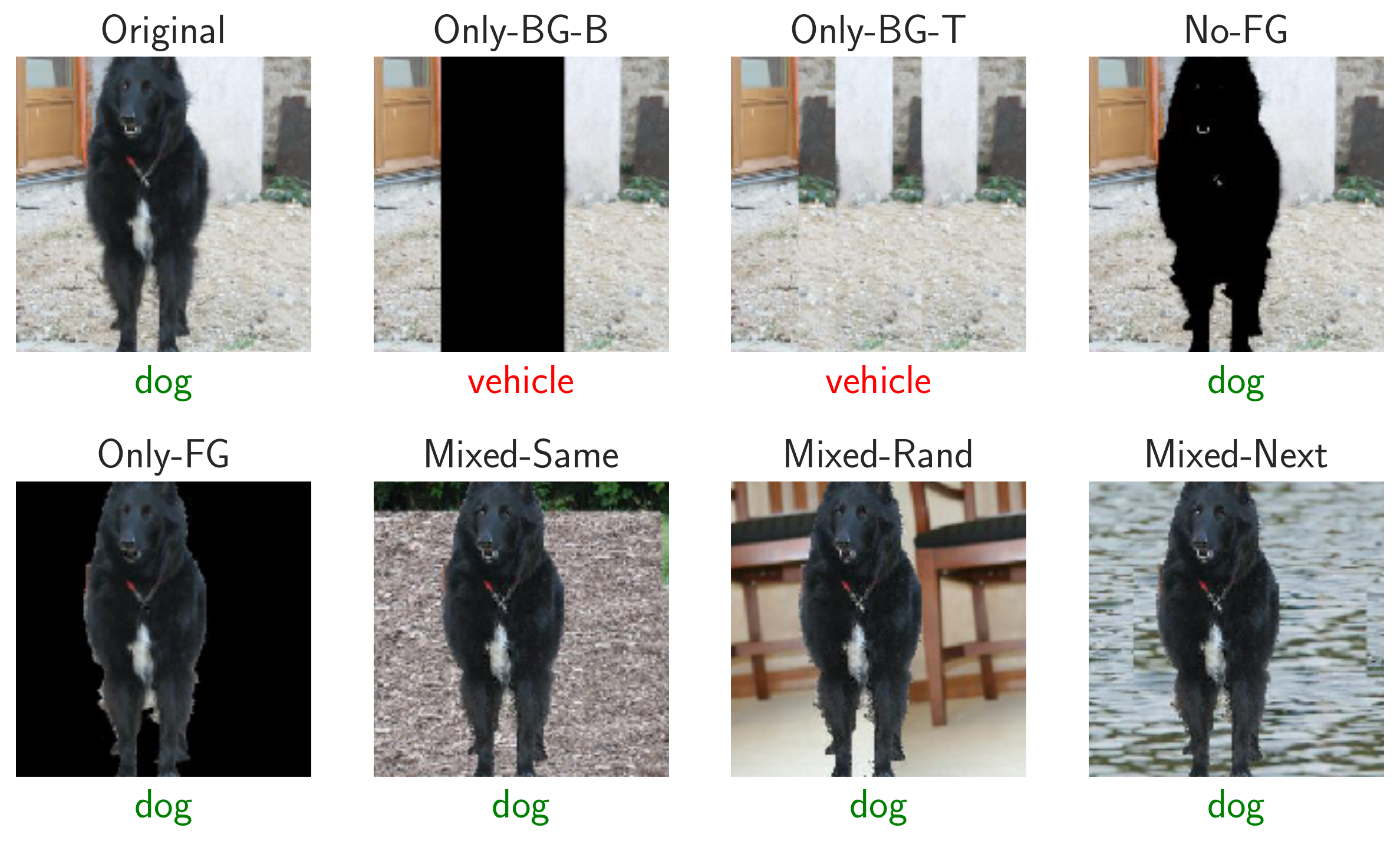}
\caption{\dsname variations---Dog.}
\label{fig:apx_dog}
\end{figure}

\begin{figure}[h]
\centering
\includegraphics[width=0.9\linewidth]{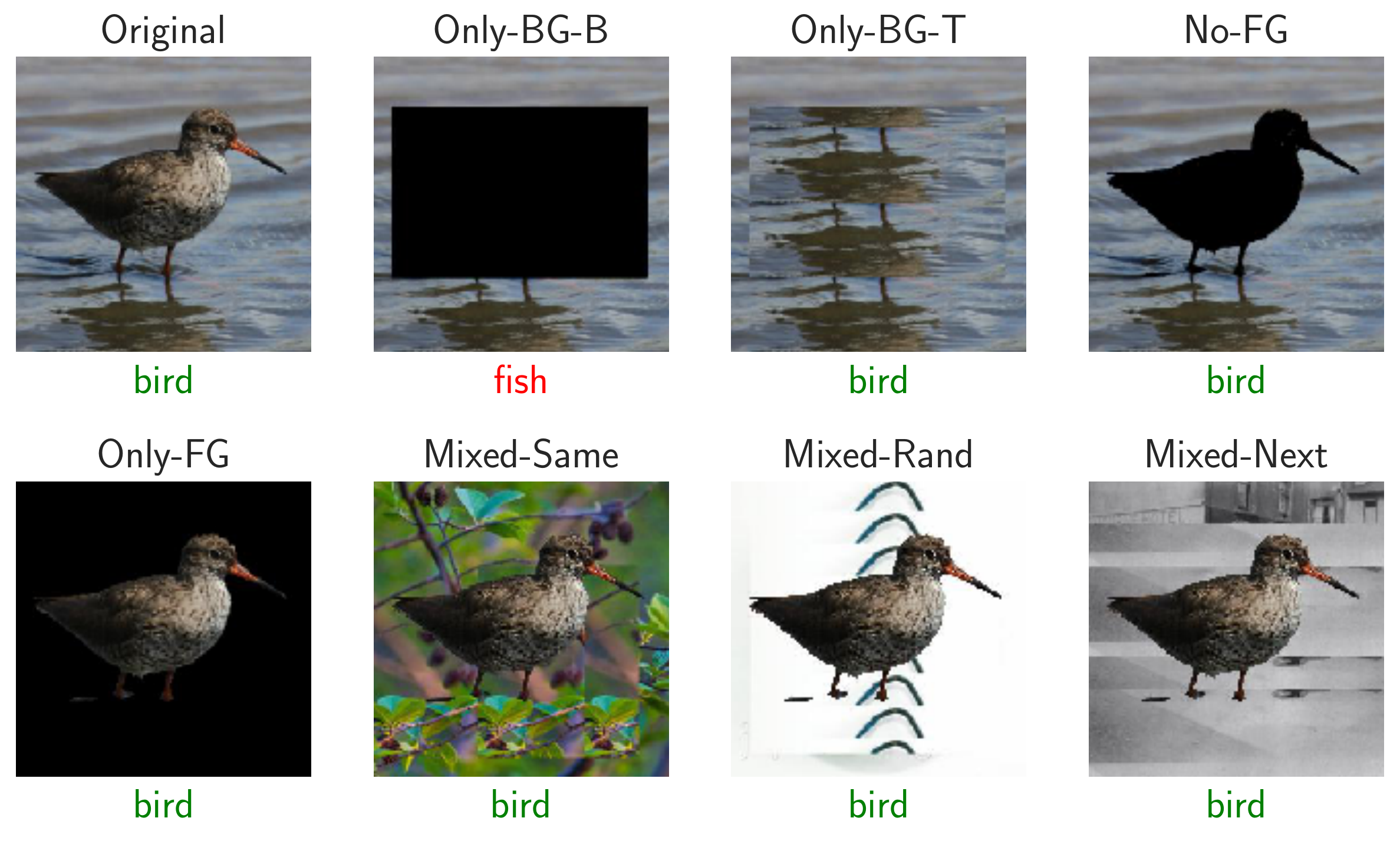}
\caption{\dsname variations---Bird.}
\label{fig:apx_bird}
\end{figure}

\begin{figure}[h]
\centering
\includegraphics[width=0.9\linewidth]{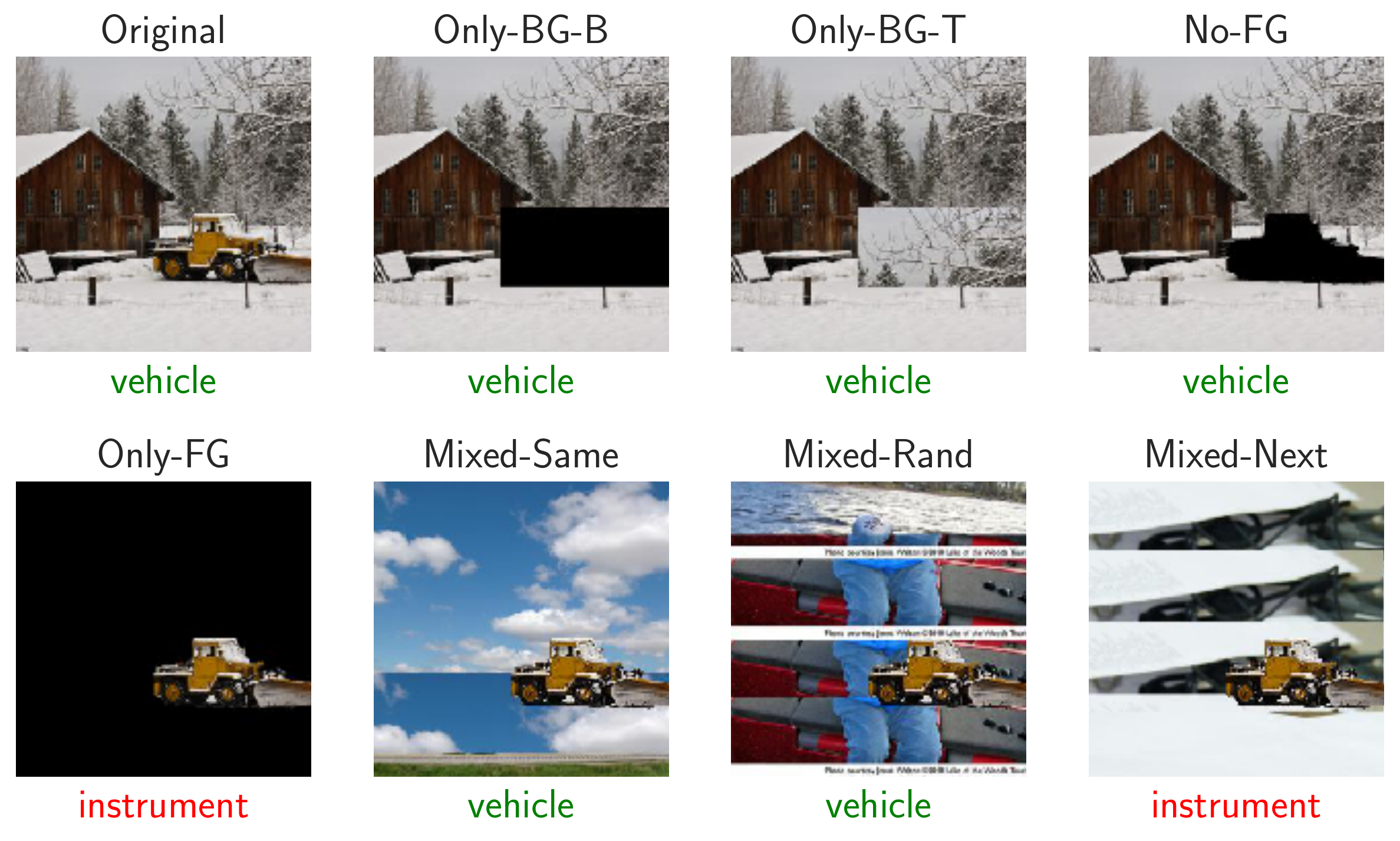}
\caption{\dsname variations---Vehicle.}
\label{fig:apx_vehicle}
\end{figure}

\begin{figure}[h]
\centering
\includegraphics[width=0.9\linewidth]{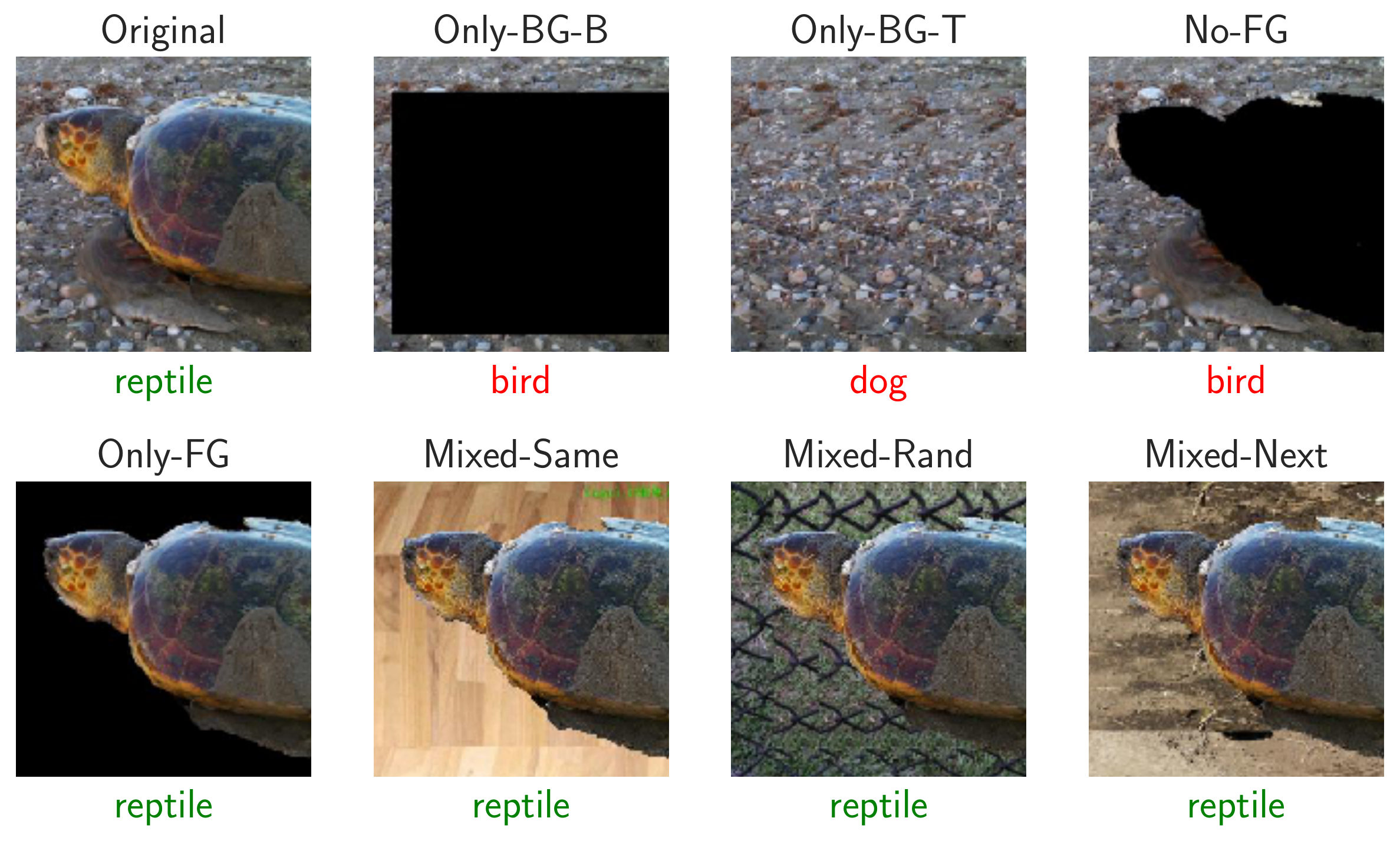}
\caption{\dsname variations---Reptile.}
\label{fig:apx_reptile}
\end{figure}

\begin{figure}[h]
\centering
\includegraphics[width=0.9\linewidth]{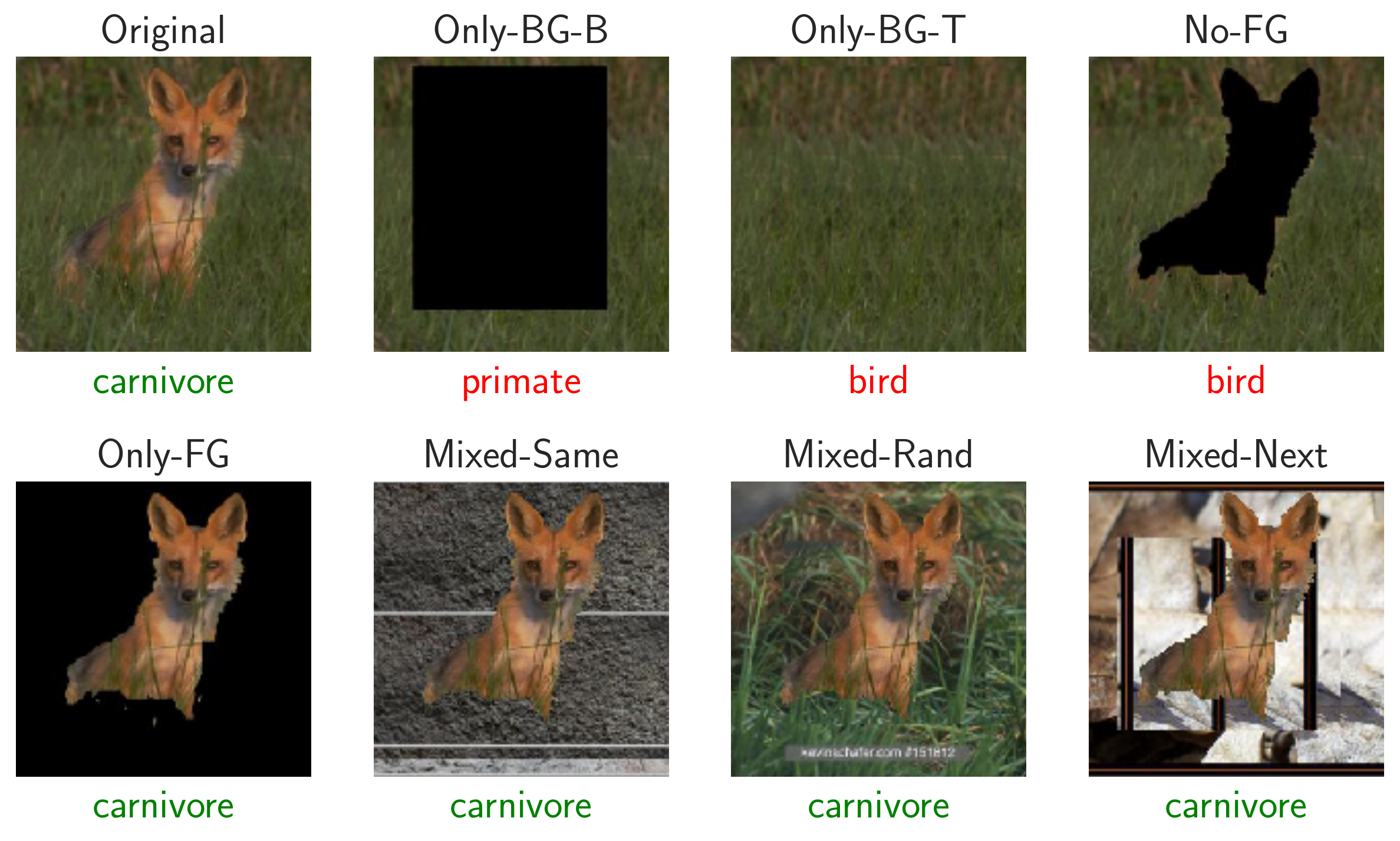}
\caption{\dsname variations---Carnivore.}
\label{fig:apx_carnivore}
\end{figure}

\begin{figure}[h]
\centering
\includegraphics[width=0.9\linewidth]{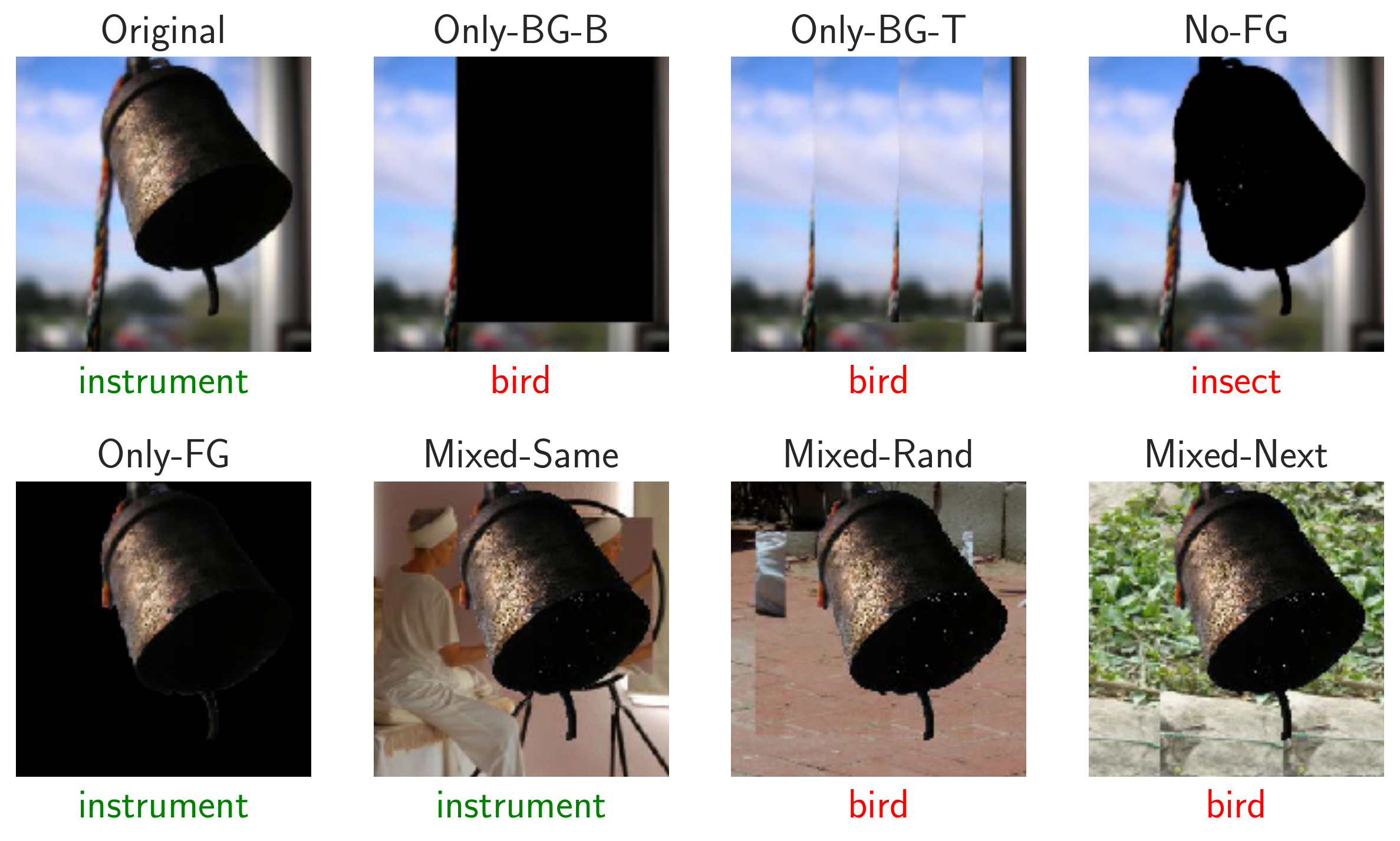}
\caption{\dsname variations---Instrument.}
\label{fig:apx_instrument}
\end{figure}

\begin{figure}[h]
\centering
\includegraphics[width=0.9\linewidth]{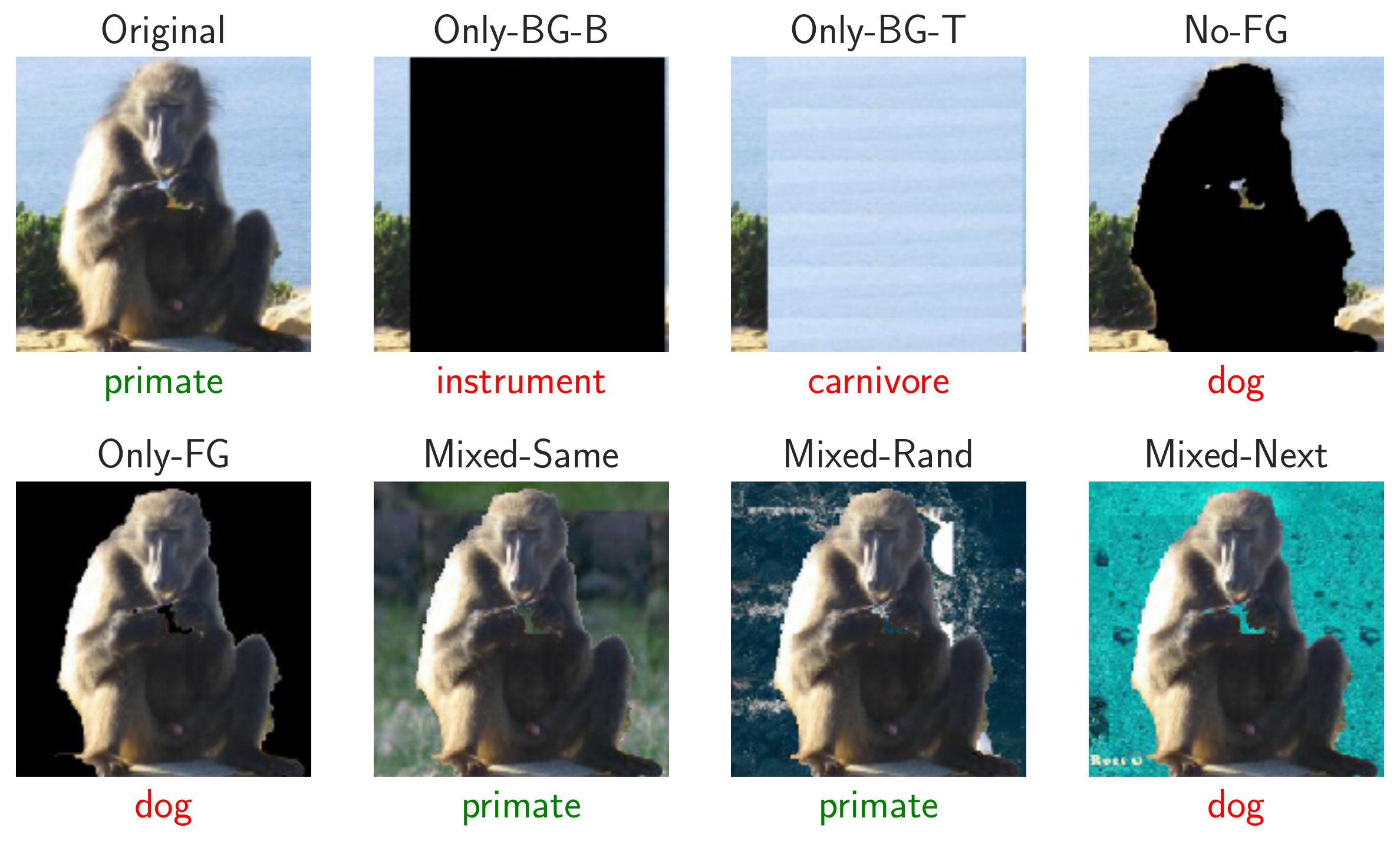}
\caption{\dsname variations---Primate.}
\label{fig:apx_primate}
\end{figure}

\begin{figure}[h]
\centering
\includegraphics[width=0.9\linewidth]{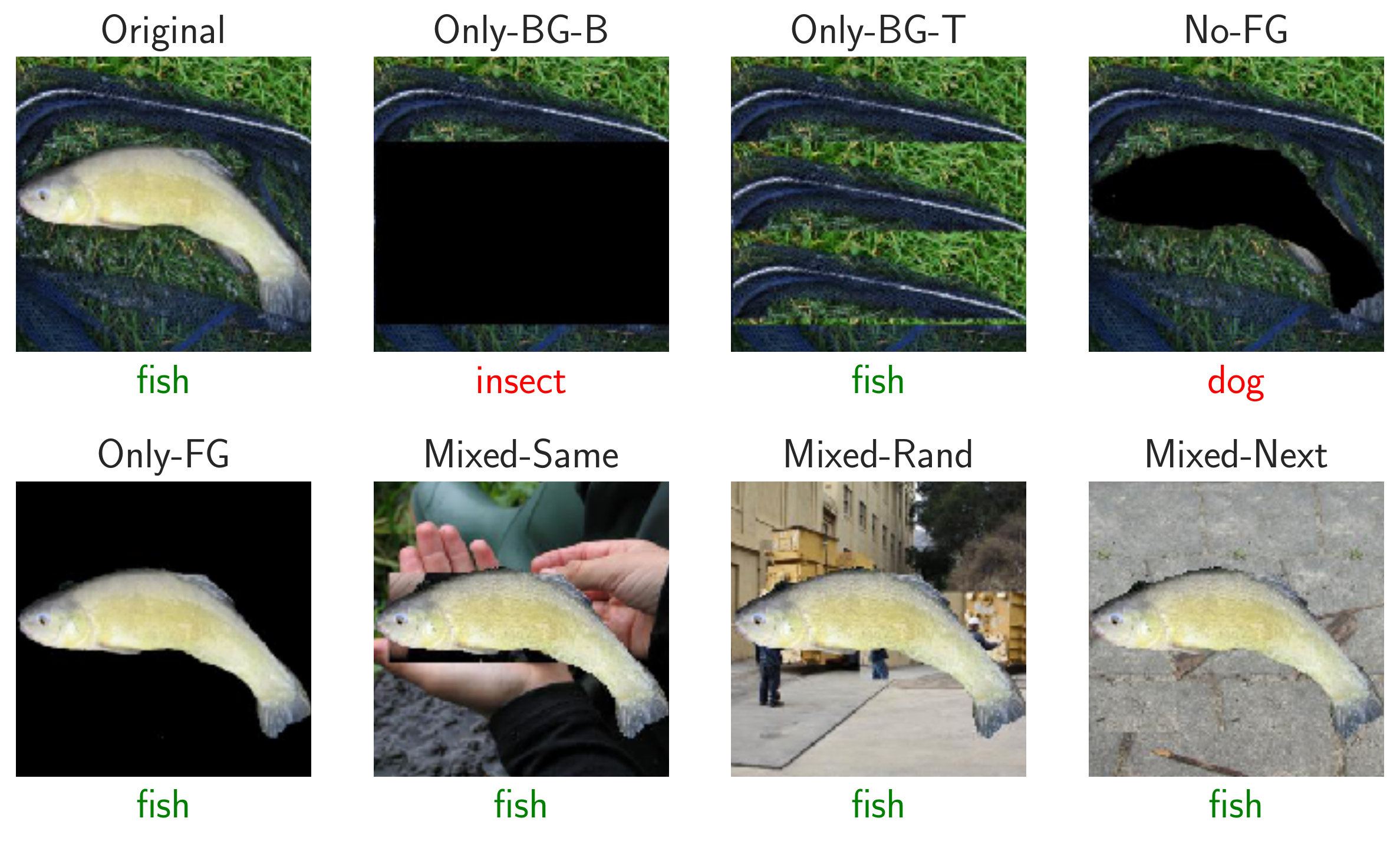}
\caption{\dsname variations---Fish.}
\label{fig:apx_fish}
\end{figure}

\clearpage

\section{Adversarial Backgrounds}
\label{appendix:adv}
We include the 5 most fooling backgrounds for all classes, the fool rate for each of those 5 backgrounds, and the total fool rate across all backgrounds from that class (on the left of each row) here.
\begin{figure}[h]
\centering
\includegraphics[width=0.9\linewidth]{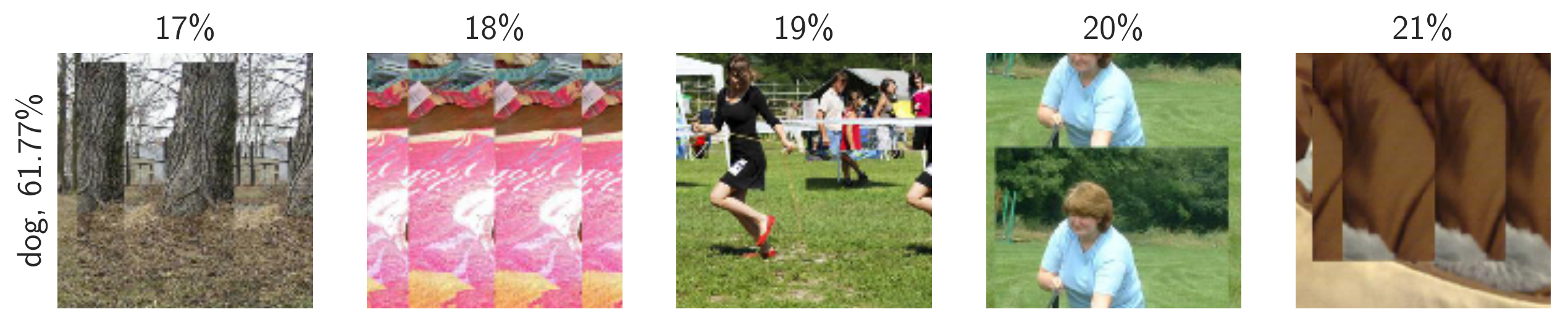}
\caption{Most adversarial backgrounds---Dog.}
\end{figure}

\begin{figure}[h]
\centering
\includegraphics[width=0.9\linewidth]{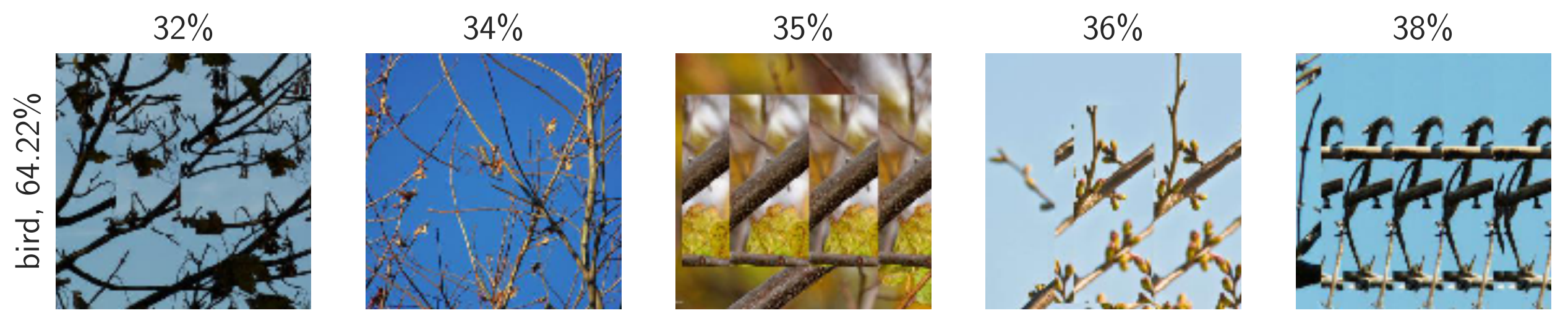}
\caption{Most adversarial backgrounds---Bird.}
\end{figure}

\begin{figure}[h]
\centering
\includegraphics[width=0.9\linewidth]{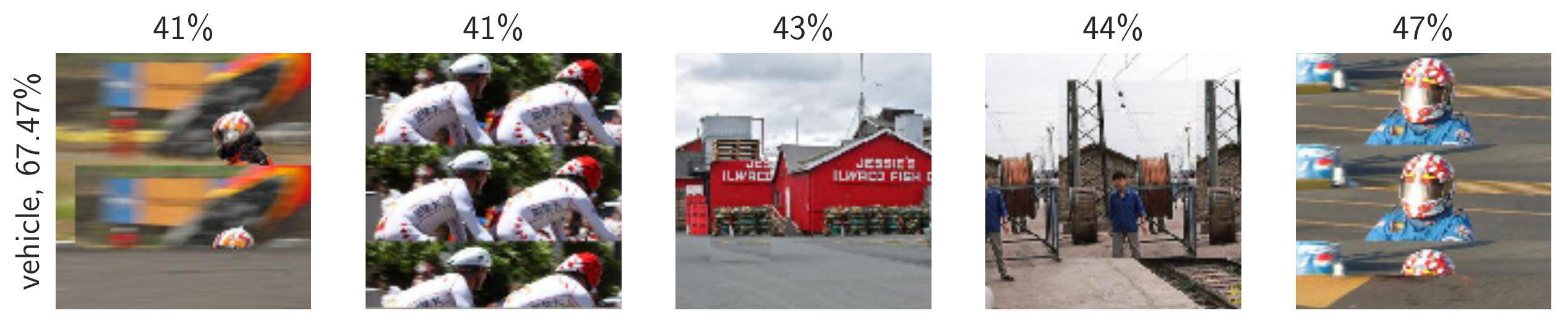}
\caption{Most adversarial backgrounds---Vehicle.}
\end{figure}

\begin{figure}[h]
\centering
\includegraphics[width=0.9\linewidth]{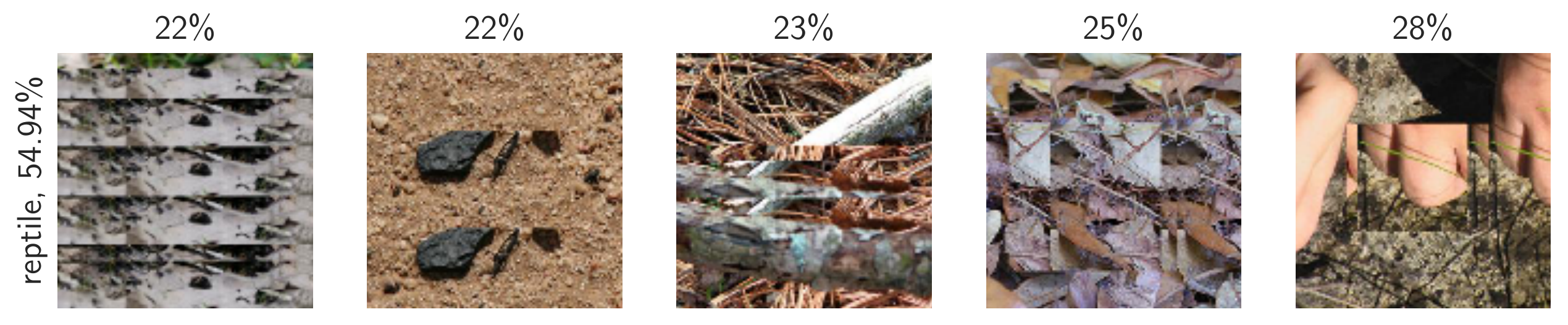}
\caption{Most adversarial backgrounds---Reptile.}
\end{figure}

\begin{figure}[h]
\centering
\includegraphics[width=0.9\linewidth]{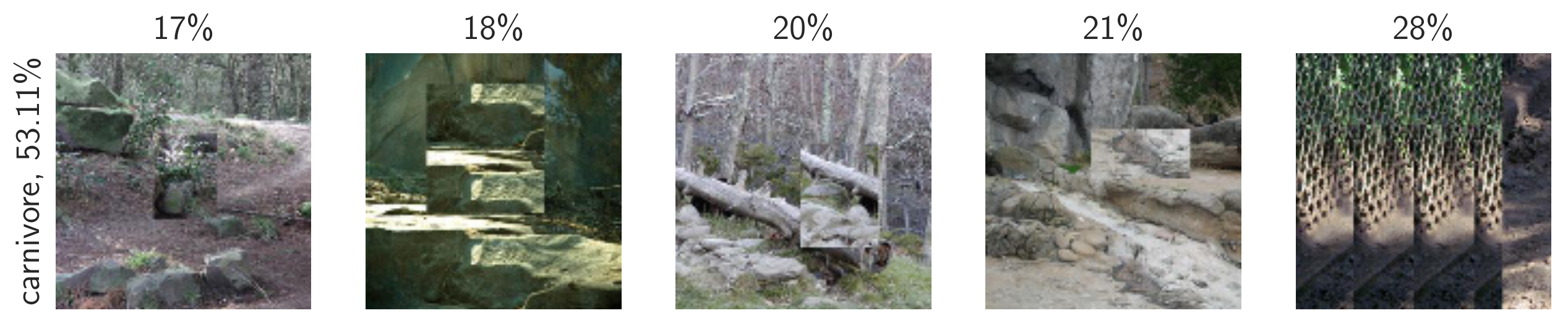}
\caption{Most adversarial backgrounds---Carnivore.}
\end{figure}

\begin{figure}[h]
\centering
\includegraphics[width=0.9\linewidth]{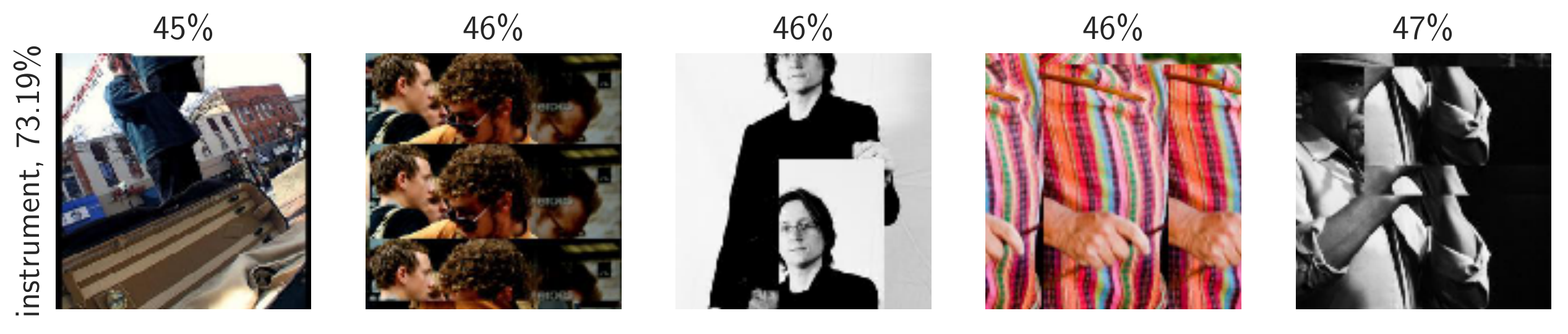}
\caption{Most adversarial backgrounds---Instrument.}
\end{figure}

\begin{figure}[h]
\centering
\includegraphics[width=0.9\linewidth]{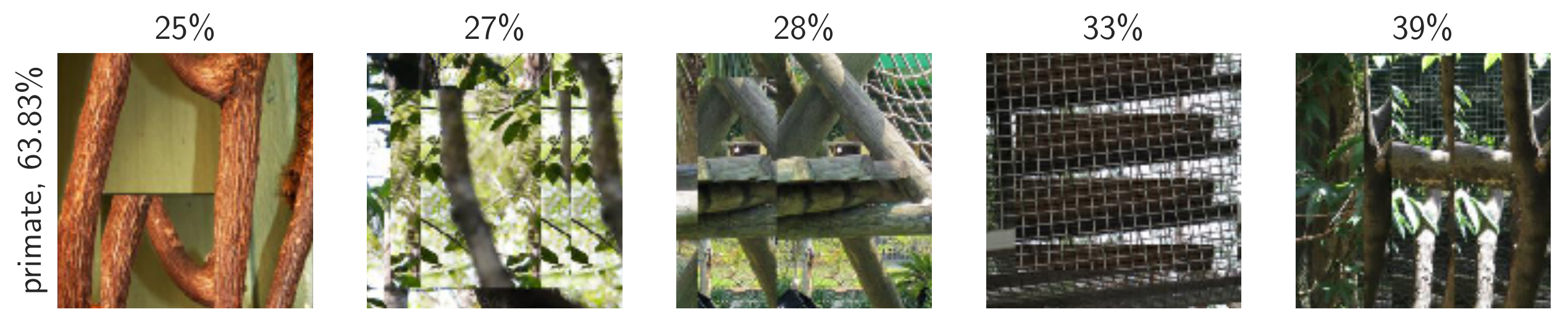}
\caption{Most adversarial backgrounds---Primate.}
\end{figure}

\begin{figure}[h]
\centering
\includegraphics[width=0.9\linewidth]{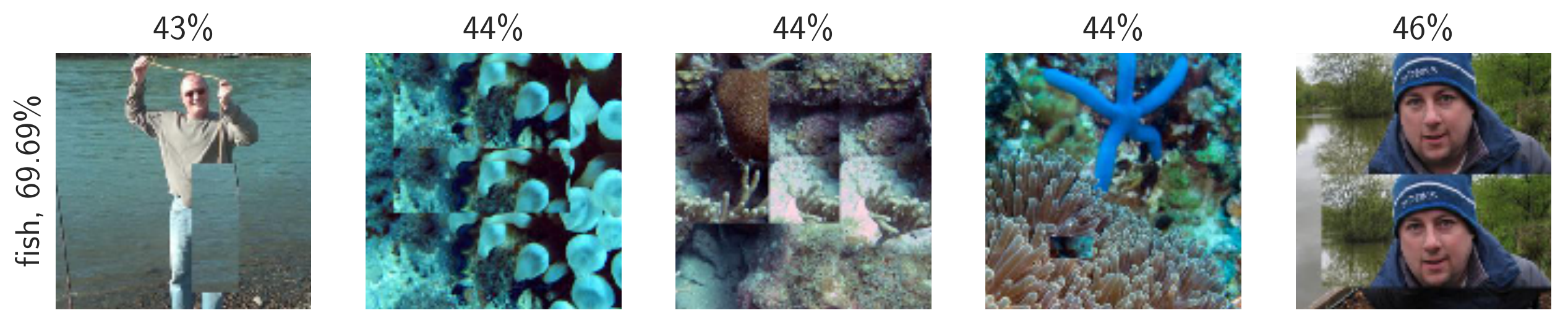}
\caption{Most adversarial backgrounds---Fish.}
\end{figure}

\clearpage

\section{Examples of Fooling Backgrounds in Unmodified Images}
\label{appendix:misleading_backgrounds}

\begin{figure}[h]
    \centering
    \includegraphics[width=\linewidth]{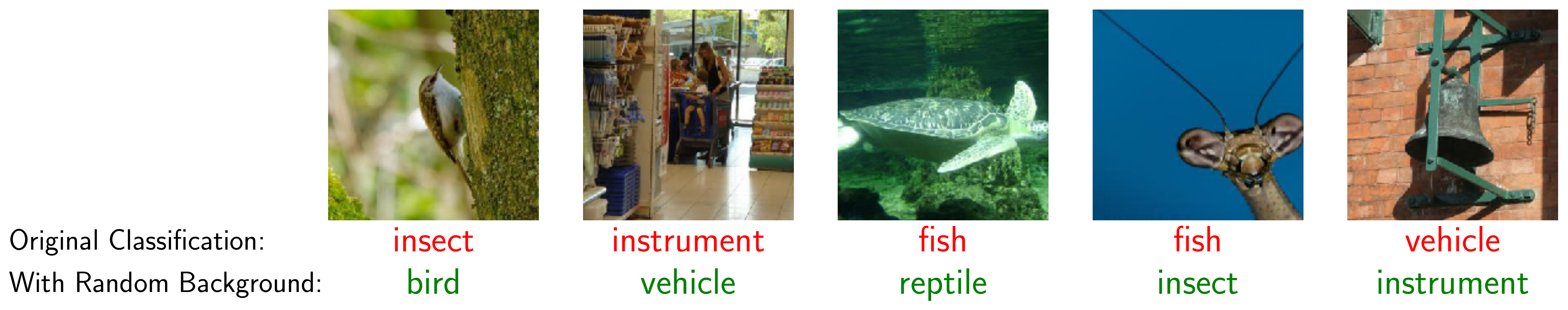}
    \caption{Images that are incorrectly classified (as the class on the top row,
      which is the same class that their background alone from \obgtds is
      classified as), but are correctly classified (as the class on the bottom row)
      when the background is randomized. Note that these images have
      confusing backgrounds that could be associated with another class.}
    \label{fig:unorthodox_bgs}
\end{figure}

We visualize examples of images where the background of the full original image
actually fools models in Figure~\ref{fig:unorthodox_bgs}. For these images, models classify the foreground alone correctly,
but they predict the same wrong class on the full image and the background. We denote these images as
``BG Fools'' in Table~\ref{table:fine_grained_defs} and Figure~\ref{fig:apx_fine_grained_bars}.
While this category is relatively rare (accounting for just 3\% of the \origds-trained model's predictions),
they reveal a subset of original images where background signal hurts classifier performance.
Qualitatively, we 
observe that these images all have confusing or misleading backgrounds.

\end{document}